\documentclass[numbers,sort&compress,preprint,12pt]{elsarticle}

\usepackage{amsmath}
\usepackage{hyperref}
\usepackage{url}
\usepackage{pifont} 
\usepackage{caption2}     
\usepackage{cancel}
\usepackage{ulem}
\usepackage{float}

\usepackage{algorithm}
\usepackage{algpseudocode}

\usepackage{graphicx}     
\usepackage{subfigure}
\graphicspath{{figures/}} 
\usepackage{booktabs} 
\usepackage{multirow}
\usepackage{natbib}
\usepackage[sectionbib]{bibunits}

\newtheorem{theorem}{\textbf{Theorem}}

\newenvironment{proof}{\par\noindent\textit{Proof.} \ignorespaces}{\hfill$\square$\par}

\usepackage{amssymb}

\usepackage{seqsplit}       

\usepackage{lmodern}
\usepackage{tikz}           
\usepackage[]{xcolor}       
\usepackage{ulem} 
\usepackage{multicol}
\usepackage{arydshln}   
\makeatletter

\usepackage{placeins}

\makeatother

\usetikzlibrary{calc,decorations.pathmorphing,decorations.pathreplacing,fadings,shadings,shapes,arrows,positioning,patterns,automata,shadows.blur,fit,arrows.meta} 

\tikzset{
	>=stealth',
	punkt/.style={
		rectangle,
		rounded corners,
		draw=black, very thick,
		text width=12em,
		minimum height=2em,
		text centered,
		fill=black!5},
	pil/.style={
		->,
		thick,
		shorten <=2pt,
		shorten >=2pt,}
}
\tikzstyle{class}=[
rectangle,
draw=black,
text centered,
anchor=north,
text=black,
text width=7cm,
shading=axis,
blur shadow={shadow blur steps=5},
bottom color=black!10,top color=white,shading angle=45]
\tikzstyle{whitebox}=[rectangle, draw=black, text centered, anchor=north, text=black, text width=7cm, bottom color=white,top color=white]
\tikzstyle{estimator} =[class, bottom color=blue!10, top color=blue!10, shading angle=45]
\tikzstyle{class2} = [class, text width=4.7cm]
\tikzstyle{class3} = [class, text width=3.2cm]
\tikzstyle{inharrow}=[->, >=open triangle 90, thick]
\tikzstyle{comp}=[->, >=diamond, thick]

\defaultbibliography{elsarticle}
\defaultbibliographystyle{elsarticle-num-names}

\journal{Neural Networks}

\begin{document}

\begin{frontmatter}

\title{Energy-based diffusion generator for efficient sampling of Boltzmann distributions} 


\author{Yan Wang\textsuperscript{a},\ \ Ling Guo\textsuperscript{b},\ \ Hao Wu\textsuperscript{*c},\ \ Tao Zhou\textsuperscript{d}} 

\affiliation{organization={School of Mathematical Sciences, Tongji University},
            city={Shanghai},
            country={China}}
\affiliation{organization={Department of Mathematics, Shanghai Normal University},
            city={Shanghai},
            country={China}}
\affiliation{organization={School of Mathematical Sciences, Institute of Natural Sciences and MOE-LSC, \\ Shanghai Jiao Tong University},
            city={Shanghai},
            country={China},
            addressline={,\ *hwu81@sjtu.edu.cn}}         
\affiliation{organization={Institute of Computational Mathematics and Scientific/Engineering
Computing, AMSS, Chinese Academy
of Sciences},
            city={Beijing},
            country={China}}



\newpage

\begin{abstract}
Sampling from Boltzmann distributions, particularly those tied to high dimensional and complex energy functions, poses a significant challenge in many fields. In this work, we present the Energy-Based Diffusion Generator (EDG), a novel approach that integrates ideas from variational autoencoders and diffusion models. EDG uses a decoder to generate Boltzmann-distributed samples from simple latent variables, and a diffusion-based encoder to estimate the Kullback-Leibler divergence to the target distribution. Notably, EDG is simulation-free, eliminating the need to solve ordinary or stochastic differential equations during training. Furthermore, by removing constraints such as bijectivity in the decoder, EDG allows for flexible network design. Through empirical evaluation, we demonstrate the superior performance of EDG across a variety of sampling tasks with complex target distributions, outperforming existing methods.
\end{abstract}

\begin{keyword}
Boltzmann distribution \sep Energy-based model \sep Generative model \sep Diffusion model \sep Variational autoencoder

\end{keyword}

\end{frontmatter}


\renewcommand{\theequation}{\arabic{equation}}
\renewcommand{\thefigure}{\arabic{figure}}
\renewcommand{\thetable}{\arabic{table}}
\setcounter{equation}{0}
\setcounter{figure}{0}
\setcounter{table}{0}
\setcounter{theorem}{0}
\thispagestyle{empty}
\section{Introduction} \label{sec:intro}

In various fields such as computational chemistry, statistical physics, and machine learning, the challenge of sampling from a Boltzmann distribution corresponding to a high-dimensional and complex energy function is ubiquitous \textcolor{black}{\cite{rapaport2004art}}. Unlike training tasks for data-driven generative models, where pre-sampled data can be utilized to learn complex distributions, sampling from Boltzmann distributions presents a unique and significant challenge due to the lack of readily available data \textcolor{black}{\cite{martino2013flexibility,rubinstein2016simulation}}.
For example, simulating the phase transition of the Ising model can be framed as a sampling problem given the energy function, which presents a complex and difficult problem that has yet not to be effectively addressed \textcolor{black}{\citep{binder2001monte, walker2020deep}}.

Markov Chain Monte Carlo (MCMC) methods \textcolor{black}{\cite{mc-1}}, along with Brownian and Hamiltonian dynamics \cite{mc-2, araki2013adaptive, araki2015efficient, mc-6}, have offered a pivotal solution to the challenge of sampling from high-dimensional distributions. These methods operate by iteratively generating candidates and updating samples, ultimately achieving asymptotic unbiasedness at the limit of infinite sampling steps. 
\textcolor{black}{Meanwhile, some enhanced sampling methods \cite{vanden2012rare, martinsson2019simulated} developed for rare event problems accelerate sampling efficiency while significantly reducing relative statistical errors, even for low-probability regions.}

In recent years, researchers have proposed adaptive MCMC as a strategy for generating candidate samples, showcasing notable advancements in augmenting the efficiency and effectiveness of the sampling process \citep{mc-4,mc-5,galliano2024policy}.
\textcolor{black}{However, designing optimal loss functions is always challenging for the sampling problems. Many distributional discrepancies used in generative models, such as Jensen-Shannon divergence, maximum mean discrepancy and Wasserstein distance, require access to samples from the true distribution, which are unavailable in our setting. Stein discrepancy derived from Stein's identity \cite{vi-8}, and the Kullback-Leibler (KL) divergence, between the distribution of the generated samples and the target distribution, may serve as potential alternatives for sampling.}


Variational inference (VI) is a widely used approach to sampling problems. It employs a generator that efficiently produces samples to approximate the target Boltzmann distribution, and optimizes the generator’s parameters using the KL divergence, without requiring true samples from the target distribution.
Due to its ability to model complex distributions and provide explicit probability density functions, normalizing flow (NF) \cite{dinh2014nice,dinh2016density,chao2023training,nijkamp2020mcmc} has been extensively applied to construct generators for VI methods \cite{key-4,vi-2}. Furthermore, a growing body of research has expanded their applicability \textcolor{black}{across diverse domains \cite{dibak2022temperature, kohler2021smooth, ising, wirnsberger2023estimating, schebek2024efficient, coretti2025learning}}.
However, the bijective nature of NFs imposes a constraint on their effective capacity, often rendering it insufficient for certain sampling tasks. \textcolor{black}{Continuous normalizing flows \cite{klein2024transferable,kohler2020equivariant,albergo2022building,jung2024normalizing}} represent a specific subclass of NF in which the invertible transformation is defined through an ordinary differential equation. For more relevant studies, please refer to the Related Work in Sec. \ref{sec:relatedwork}.

With the prosperity of \textcolor{black}{diffusion based generative models \citep{diff-2,key-2,key-3, bou2014metropolis}}, they have been applied to address challenges in the sampling problem. By training time-dependent score matching neural networks, methods proposed in \citep{diff-6,diff-4,diff-3} shape the Gaussian distribution into complex target densities, employing the KL divergence as the loss function. \textcolor{black}{\cite{diff-1} also explores multiple connections between diffusion and other VI methods.} To mitigate mode-seeking issues, \citep{diff-7} introduces the log-variance loss, showcasing favorable properties. Additionally, an alternative training objective is outlined in \citep{diff-8}, relying on flexible interpolations of the energy function and demonstrating substantial improvements for multi-modal targets. However, a common drawback of these methods is their reliance on numerical differential equation solvers for computing time integrals, which can lead to substantial computational costs. 

In this research endeavor, we present a novel approach termed the energy-based diffusion generator (EDG), drawing inspiration from both the variational autoencoder (VAE) technique  \citep{kingma2019introduction} and the diffusion model. The architecture of EDG closely resembles that of VAE, comprising a decoder and an encoder. The decoder is flexible in mapping latent variables distributed according to tractable distributions to samples, without the imposition of constraints such as bijectivity, and we design in this work a decoder based on generalized Hamiltonian dynamics to enhance sampling efficiency. The encoder utilizes a diffusion process, enabling the application of score matching techniques for precise and efficient modeling of the conditional distribution of latent variables given samples. Unlike existing diffusion-based sampling methods, the loss function of EDG facilitates the convenient computation of unbiased estimates in a random mini-batch manner, removing the need for numerical solutions to ordinary differential equations (ODEs) or stochastic differential equations (SDEs) during training. Numerical experiments conclusively demonstrate the effectiveness of EDG.

\section{Related work and preliminaries}
\subsection{\textcolor{black}{Related work}}\label{sec:relatedwork}
\textcolor{black}{\textbf{Variational inference-based sampling algorithms:} The Boltzmann generator (BG) \cite{bg3, vi-2} is one of the most representative sampling methods based on VI. It leverages NFs to parameterize a trainable and analytically tractable density, with parameter optimization carried out by minimizing the KL divergence between the surrogate and target distributions.
Here, NFs are composed of stacked bijective transformations, allowing the density to be computed in closed form. In addition, the combination of MCMC and VI methods stands as a current focal point in research \citep{vi-3,vi-4,vi-6,vi-5,shen2021randomizing}. This combination seeks to harness the strengths of both approaches, offering a promising avenue for addressing the challenges associated with sampling from high-dimensional distributions and enhancing the efficiency of probabilistic modeling.}

\textcolor{black}{\textbf{Adaptive MCMC:} Recent work has augmented MCMC sampling with nonlocal transition kernels parameterized by neural networks, including the NF-based method proposed by Vanden-Eijnden et al.~\cite{gabrie2022adaptive} and the diffusion model-based method \cite{akhound2024iterated}. This synergistic approach enables neural networks to accelerate MCMC sampling while the model are trained in a data-driven mode from samples of MCMC.}

\textcolor{black}{\textbf{Stein-based models:} Designing optimal loss functions for sampling problems is always challenging for the sampling problems. 
Stein discrepancy has emerged as a promising candidate \cite{vi-7, vi-8, grathwohl2020learning, di2021neural}, as it allows direct evaluation of the discrepancy between generated samples and the target distribution, without requiring real samples from the target or access to the density of the generated distribution. In particular, the kernelized Stein discrepancy \cite{vi-8} offers a practical and convenient criterion, which can test the goodness of fit easily.
The trainable Stein Discrepancy \cite{grathwohl2020learning,di2021neural} further leverages neural networks to parameterize a class of functions based on the Stein’s Identity, using the learned discrepancy to enhance its capability.}

\textcolor{black}{\textbf{Latent diffusion models:} Latent diffusion models, which are primarily designed to address generative modeling in data-driven scenarios, have recently gained significant attention \cite{rombach2022high, fu2024latent, zheng2024ld}. Their main idea to use a pre-trained encoder and decoder to obtain a latent space that both effectively represents the data and facilitates efficient sampling, with the diffusion model learning the distribution of latent variables. However, its integration with sampling problems remains to be explored.}

\subsection{Preliminaries and Setup} \label{sec:setup}
In this work, we delve into the task of crafting generative models for the purpose of sampling from the Boltzmann distribution driven by a predefined energy $U:\mathbb R^d\to\mathbb R$:
\[
\pi(x)=\frac{1}{Z}\exp(-U(x)),
\]
where the normalizing constant $Z = \int \exp(-U(x))\mathrm dx$ is usually computationally intractable.
To tackle this challenge, the BG \citep{key-4}, along with its various extensions \citep{bg2, bg3, bg4}, has emerged as a prominent technique in recent years.
In this work, we aim to overcome the limitations of BG by employing a generator with fewer structural constraints (e.g., without requiring bijection), and design a flexible model tailored for sampling tasks.

Our focus now shifts to a generator akin to the VAE. This generator produces samples by a decoder as
\begin{equation*}
x|z_0 \sim p_D(x|z_0;\phi),
\end{equation*}
where $z_0 \sim p_D(z_0)$ is a latent variable drawn from a known prior distribution, typically a standard multivariate normal distribution. The parameter $\phi$ characterizes the decoder, and we define $p_D(x|z_0;\phi)$ as a Gaussian distribution $\mathcal{N}(x|\mu(z_0;\phi), \Sigma(z_0;\phi))$, with both $\mu$ and $\Sigma$ parameterized by neural networks (NNs). Similar to the VAE, we aim to train the networks $\mu$ and $\Sigma$ such that the marginal distribution $p_D(x)$ of the generated samples aligns with the target distribution.


It is important to note that, unlike conventional data-driven VAEs, we do not have access to samples from the target distribution $\pi(x)$. In fact, obtaining such samples is precisely the goal of the generator. As a result, the variational approximation of the KL divergence $D_{\mathrm{KL}}\left(\pi(x) || p_D(x)\right)$ cannot be used to train the model. Instead, in this work, we consider the divergence $D_{\mathrm{KL}}\left(p_{D}(x)||\pi(x)\right)$ and its upper bound:
\textcolor{black}{\begin{align}
D_{\mathrm{KL}}\left(p_{D}(x)||\pi(x)\right) & \le D_{\mathrm{KL}}\left(p_{D}(z_{0}) p_{D}(x|z_{0};\phi)||\pi(x) p_{E}(z_{0}|x;\theta)\right)\nonumber \\
 & = \mathbb{E}_{p_{D}(z_{0}) p_{D}(x|z_{0};\phi)}\left[\log\frac{p_{D}(z_{0})p_{D}(x|z_{0};\phi)}{p_{E}(z_{0}|x;\theta)}+U(x)\right]\nonumber \\
 & \qquad +\log Z. \label{eq:vae-loss}
\end{align}}
Here, the parametric distribution $p_E(z_0|x;\theta)$ defines an encoder that maps from $x$ to the latent variable $z_0$, and the equality is achieved if $p_E(z_0|x;\theta)$ matches the conditional distribution of $z$ for a given $x$, deduced from the decoder. \textcolor{black}{The detailed proof is provided in \ref{append:proof-vae-loss}}.

It seems that we have only increased the complexity of the problem, as we are still required to approximate a conditional distribution. However, in the upcoming section, we demonstrate that we can effectively construct the encoder using the diffusion model \citep{diff-1, key-2} and optimize all parameters without the need to numerically solve ordinary or stochastic differential equations.

\textcolor{black}{\section{Energy-based diffusion generator}}
The diffusion model \citep{key-2,key-3} has emerged in recent years as a highly effective approach for estimating data distributions. Its core idea is to construct a diffusion process that progressively transforms data into simple white noise, and learn the reverse process to recover the data distribution from noise. In this work, we apply the principles of the diffusion model by incorporating a diffusion process into the latent space, enabling us to efficiently overcome the challenges arising from the variational framework for the sampling problem defined by Eq. \eqref{eq:vae-loss}. We refer to the model produced by this method as the energy-based diffusion generator (EDG).

~

\subsection{Model architecture}\label{sec:model}

In the EDG framework, we initiate a diffusion process from the latent variable $z_0$ and combine it with the decoder, resulting in what we term the ``decoding process'' $p_D$ 
\begin{eqnarray}
&&z_{0}\in \mathbb R^d \sim p_{D}(z_{0})\triangleq \mathcal{N}(x|0,I), \quad x|z_{0} \sim p_{D}(x|z_{0};\phi) \nonumber\\
&&\mathrm{d}z_{t} = f(z_{t},t)\mathrm{d}t+g(t)\mathrm{d}\textcolor{black}{W_{t}}, \quad t\in[0, T] \label{eq:forward-sde}
\end{eqnarray}
where $W_{t}$ is the standard Wiener process, $f(\cdot ,t): \mathbb R^{d}\rightarrow \mathbb R^{d}$ acts as the drift coefficient, and $g(\cdot): R\rightarrow R$ serves as the diffusion coefficient. To streamline notation, we denote the probability distribution defined by the decoding process as $p_D$,
\textcolor{black}{and represent the variables defined by the SDE as $z_{[\cdot]}=\{z_t\}_{t \in [0, T]}$.}
In typical SDEs applied in diffusion models, two critical conditions hold: (a) the transition density $p_D(z_t|z_0)$ can be analytically computed without numerically solving the Fokker-Planck equation, and (b) $z_T$ is approximately uninformative with $p_D(z_T)\approx p_D(z_T|z_0)$.

If we only consider the statistical properties of the latent diffusion process, it is uninformative and only describes a transition from one simple noise to another. However, when we account for the conditional distribution of \(z_t\) given a sample \(x\), the process \(z_{[\cdot]}\) represents the gradual transformation of the complex conditional distribution \(p_D(z_0|x) \propto p_D(z_0) \cdot p_D(x|z_0)\) into the tractable distribution \(p_D(z_T|x) = p_D(z_T)\), where the independence between $z_T$ and $x$ results from the independence between $z_0$ and $z_T$ (see \ref{append:independence}). This implies that, starting from \(z_T\sim p_D(z_T)\), we can obtain samples from \(p_D(z_0|x)\) by simulating the following reverse-time diffusion equation \citep{diff-9}:
\begin{equation}\label{eq:reverse-sde}
\mathrm{d}z_{\tilde{t}}=-\left(f(z_{\tilde{t}},\tilde{t})-g(\tilde{t})^2\nabla_{z_{\tilde t}}\log p_{D}(z_{\tilde{t}}|x)\right)\mathrm{d}\tilde{t}+g(\tilde{t})\mathrm{d}\textcolor{black}{\widetilde{W}_{\tilde{t}}},    
\end{equation}
where $\tilde t=T-t$ denotes the reverse time. \textcolor{black}{The process $\widetilde{W}$ is a distinct standard Wiener process, whose non-trivial relationship with $W$ is formally discussed in \cite{diff-9}.}
As in conventional diffusion models, practical implementation of this simulation is challenging due to the intractability of the score function \(\nabla_{z_{\tilde t}} \log p_D(z_{\tilde{t}}|x)\), and we therefore also use a neural network to approximate the score function, denoted as \(s(z_{\tilde t}, x, \tilde t; \theta)\).
This approximation leads to what we refer to as the ``encoding process'', achieved by integrating the parametric reverse-time diffusion process and the target distribution of $x$:
\begin{eqnarray}
&& x \sim \pi(x), \quad z_{T} \sim p_{E}(z_{T})\triangleq p_{D}(z_{T}) \nonumber \\
&& \mathrm{d}z_{\tilde{t}}=-\left(f(z_{\tilde{t}},\tilde{t})-g(\tilde{t})^2s(z_{\tilde t},x,\tilde t;\theta)\right)\mathrm{d}\tilde{t}+g(\tilde{t})\mathrm{d}\textcolor{black}{\widetilde{W}_{\tilde{t}}},\quad \tilde{t}=T-t.\label{eq:reverse-diffusion}
\end{eqnarray}
For simplicity of notation, we refer to the distribution defined by the encoding process as $p_E$ in this paper.
\begin{figure}
\begin{center}
\includegraphics[width=1.\textwidth]{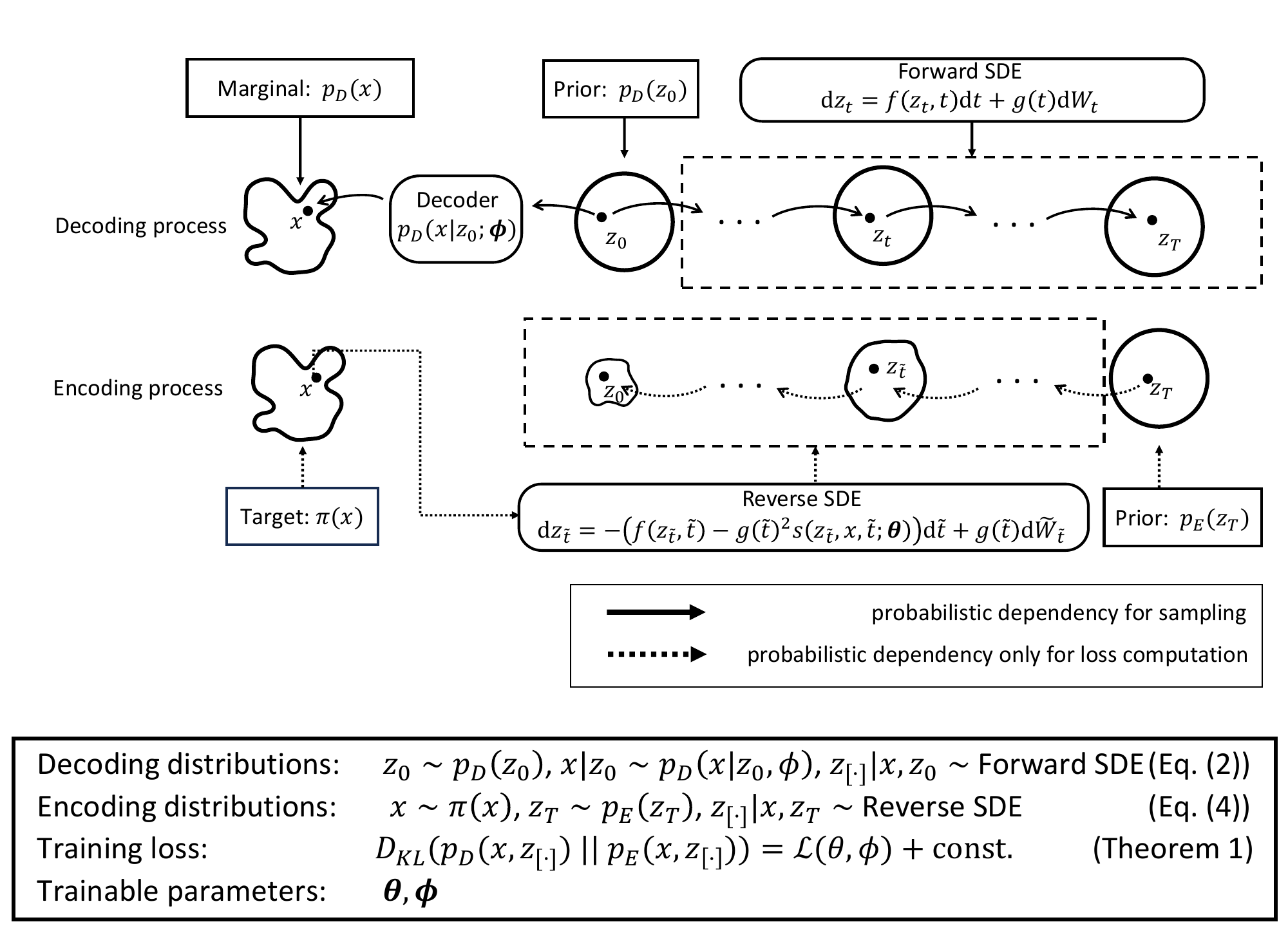}
\end{center}
\vspace{-7mm}
\caption{Illustration of the decoding and encoding processes in EDG. 
In the \textbf{decoding process}, the latent variable $z_0$ is drawn from a tractable prior $p_D(z_0)$, and a sample $x$ is generated via the decoder distribution $p_D(x | z_0)$ (modeled as Gaussian in this work). 
Simultaneously, a latent path $z_{[\cdot]} = \{z_t\}_{t \in [0, T]}$ is generated by a forward SDE.
\textcolor{black}{In the \textbf{encoding process}, \( x \sim \pi(x) \) and \( z_T \sim p_E(z_T) \) are assumed to be independently distributed, and a reverse-time SDE models the conditional distribution of the latent path given \( x \) and \( z_T \), where $s(z_{\tilde t}, x, \tilde t; \theta)$ approximates the score function $\nabla_{z_{\tilde t}} \log p_D(z_{\tilde t} | x;\phi)$. 
Given that training samples of \( x \) and \( z_t \) are generated exclusively by the decoding process, while the encoding process is used solely for loss computation (not for actual sampling), we use \textbf{solid arrows} to indicate probabilistic dependencies in the decoding process and \textbf{dashed arrows} for those in the encoding process.} Both the decoder parameters (mean and covariance) and the score model are parameterized by neural networks and jointly optimized via the loss function $\mathcal{L}(\theta, \phi)$.}
\label{fig:structure-decoding-encoding}
\end{figure}

\textcolor{black}{Fig.~\ref{fig:structure-decoding-encoding} provides a schematic overview of the decoding and encoding processes in EDG. The decoding process formulated in Eq.~\eqref{eq:forward-sde} can be readily simulated.
In contrast, the encoding process defined by Eq.~\eqref{eq:reverse-diffusion} is not directly simulatable, as no samples from the target distribution are available prior to training.
However, as will be discussed in Sec.~\ref{sec:loss}, by optimizing the parameters of the encoding process, we can leverage samples from the decoding process and the probabilistic model induced by the encoding process to obtain an accurate variational estimate of the KL divergence between the marginal distribution of 
$x$ produced by the decoder and the target distribution $\pi(x)$. Based on this estimate, we can jointly train both the decoder and the conditional score network in the encoding process, thereby enabling the generation of high-quality samples that closely match the target distribution. Accordingly, the encoding process in EDG functions as a powerful encoder, replacing the conventional parametric encoder typically used in variational methods (see \( p_E(z_0 \mid x; \theta) \) in Eq.~\eqref{eq:vae-loss}).}

\textcolor{black}{\subsection{Loss function}\label{sec:loss}}
\textcolor{black}{Based on the architecture of the EDG, we establish the following theorem that provides a theoretical foundation for model training. It shows that the KL divergence between the joint distributions of $(x, z_{[\cdot]})$ induced by the two processes provides an upper bound for $D_{\mathrm{KL}}(p_D(x)||\pi(x))$. For a fixed decoder $p_D(x| z_0; \phi)$, minimizing the joint divergence with respect to the score model parameters $\theta$ tightens this upper bound. Moreover, jointly optimizing both $\phi$ and $\theta$ based on the joint divergence can effectively reduce the gap between the generated and target distributions of $x$.} 

\begin{theorem}\label{thm:upperbound}
\textcolor{black}{
For the decoding and encoding processes defined by Eq.~\eqref{eq:forward-sde} and Eq.~\eqref{eq:reverse-diffusion}, we have
\begin{eqnarray}
D_{\mathrm{KL}}\left(p_{D}(x)\,\|\,\pi(x)\right) & \leq & D_{\mathrm{KL}}\left(p_{D}(x,z_{[\cdot]})\,\|\,p_{E}(x,z_{[\cdot]})\right)\label{eq:edg-inequality-1}\\
& = & \mathcal{L}(\theta,\phi) + \int_0^T \frac{g(t)^2}{2}\mathbb E_{p_D}\left[\Vert \nabla_{z_t}\log p_D(z_t)\Vert^2\right]\mathrm dt\nonumber\\
& &+ \log Z,\label{eq:edg-inequality-2}
\end{eqnarray}
where $\mathcal{L}(\theta,\phi)$ is given by
\begin{eqnarray}
\mathcal{L}(\theta,\phi) & = & \mathbb{E}_{p_{D}}\left[\log p_{D}(x|z_{0};\phi) + U(x)\right]\nonumber \\
&  & + \int_{0}^{T} \frac{g(t)^{2}}{2} \mathbb{E}_{p_{D}} \Bigg[\left\Vert s\left(z_{t},x,t;\theta\right)\right\Vert^{2} + 2\nabla_{z_{t}}\cdot s\left(z_{t},x,t;\theta\right)\Bigg]\mathrm dt.\label{eq:loss}
\end{eqnarray}
Moreover, the equality in Eq.~\eqref{eq:edg-inequality-1} holds if $z_0$ is independent of $z_T$ in the decoding process, and $s(z,x,t;\theta) \equiv \nabla_{z_t} \log p_{D}(z_t | x;\phi)$.
}
\end{theorem}
\begin{proof}
    See \ref{append:upperbound}.
\end{proof}

\vspace{3mm}
\textcolor{black}{
Since the last two terms in Eq.~\eqref{eq:edg-inequality-2} are independent of the model parameters $\theta$ and $\phi$, we can adopt $\mathcal{L}(\theta,\phi)$ in Eq.~\eqref{eq:loss} as the training loss. To improve computational efficiency, we estimate the divergence term using the Hutchinson estimator and evaluate the time integral via Monte Carlo sampling. This leads to an equivalent expression of $\mathcal{L}(\theta,\phi)$ that allows efficient and unbiased estimation:}
\begin{eqnarray}
\mathcal{L}(\theta,\phi) & = & \mathbb{E}_{p_{D}}\left[\log p_{D}(x|z_{0};\phi)+U(x)\right]\nonumber\\
 &  & +\mathbb{E}_{t\sim\mathcal U[0,T],\epsilon\sim p(\epsilon),(x,z_t)\sim p_{D}(x,z_{t})}\left[\mathcal{L}_{t}(x,z_{t},\epsilon;\theta)\right],\label{eq:loss-sa}
\end{eqnarray}
\textcolor{black}{where}
\begin{equation}\label{eq:loss-t}
\mathcal{L}_{t}(x,z_{t},\epsilon;\theta) = \frac{Tg(t)^2}{2}\left(\left\Vert s\left(z_{t},x,t;\theta\right)\right\Vert ^{2}+2\frac{\partial\left[\epsilon^{\top}s\left(z_{t},x,t;\theta\right)\right]}{\partial z_{t}}\epsilon\right),
\end{equation}
\textcolor{black}{$\mathcal{U}[0,T]$ denotes a uniform distribution over $[0,T]$, and $p(\epsilon) \sim \mathrm{Rademacher}^d$ denotes a random vector in $\mathbb{R}^d$ with i.i.d.~Rademacher entries, satisfying $\mathbb{E}[\epsilon] = 0$ and $\mathrm{Cov}(\epsilon) = I$.
This expression enables the use of stochastic gradient descent for parameter optimization. (See \ref{append:proof-loss-sa} for the proof of equivalence between Eq.~\eqref{eq:loss-sa} and Eq.~\eqref{eq:loss}.)} \textcolor{black}{The training procedure is outlined in Algorithm~\ref{alg:training}. It is worth noting that when the SDE in the decoding process follows a standard formulation used in diffusion models (see Eq.~\eqref{eq:forward-sde}), the conditional distribution $p_D(z_t| z_0)$ can be sampled exactly, and numerical integration of SDEs is not required during training.}

\begin{algorithm}[H]
    \caption{\textcolor{black}{Training procedure}}\label{alg:training}
    \begin{algorithmic}
        \For{each minibatch}
            \For{$i = 1, \ldots, \text{batch size}$}
                \State Sample $z_0^i \sim p_{D}(z_{0}^i)$ and $x^i \sim p_{D}(x^i | z_{0}^i; \phi)$;
                \State Sample $t \sim \mathcal{U}[0, T]$, $\epsilon^i \sim \text{Rademacher}^d$ and $z_t^i \sim p_{D}(z_t^i | z_{0}^i)$;
                \State Evaluate $\mathcal{L}_t(x^i, z_t^i, \epsilon^i; \theta)$ using Eq.~\eqref{eq:loss-t};
                \State Compute $L_i = \log p_D(x^i | z_0^i; \phi) + U(x^i) + \mathcal{L}_t(x^i, z_t^i, \epsilon^i; \theta)$;
            \EndFor
            \State Compute minibatch loss: $\hat{\mathcal{L}}(\theta, \phi) = \text{average of }\{L_1, L_2, \ldots\}$;
            \State Update parameters: $\phi, \theta \leftarrow \text{Update}(\nabla_{\phi, \theta} \hat{\mathcal{L}}(\theta, \phi))$;
        \EndFor
    \end{algorithmic}
\end{algorithm}



\subsection{Sample reweighting}\label{sec:marginal-encoder-density}
After training, we can use the decoder \(p_D(x|z_0)\) to generate samples and compute various statistics of the target distribution \(\pi(x)\). For example, for a quantity of interest \(O: \mathbb{R}^d \to \mathbb{R}\), we can draw \(N\) augmented samples \(\{(x^n, z_0^n)\}_{n=1}^N\) from \(p_D(z_0)p_D(x|z_0)\) and estimate the expectation \(\mathbb{E}_{\pi(x)}[O(x)]\) as follows:
\[
\mathbb{E}_{\pi(x)}[O(x)] \approx \frac{1}{N}\sum_n O(x^n).
\]
However, due to model errors, this estimation can be systematically biased. To address this, we can apply importance sampling, using \(p_D(x,z_0)=p_D(z_0)p_D(x|z_0)\) as the proposal distribution and \(p_E(x,z_0)=\pi(x)p_E(z_0|x)\) as the augmented target distribution. We can then assign each sample $(x,z_0)$ generated by the decoder an unnormalized weight:
\begin{equation}\label{eq:w}
w(x,z_{0})=\frac{\exp(-U(x))p_{E}(z_{0}|x)}{p_{D}(z_{0})p_{D}(x|z_{0})}\propto\frac{\pi(x)p_{E}(z_{0}|x)}{p_{D}(z_{0})p_{D}(x|z_{0})}
\end{equation}
and obtain a consistent estimate of \(\mathbb{E}_{\pi(x)}[O(x)]\) as:
\begin{equation}\label{eq:weight}
\mathbb{E}_{\pi(x)}[O(x)]\approx\frac{\sum_{n}w(x^{n},z_{0}^{n})O(x^{n})}{\sum_{n}w(x^{n},z_{0}^{n})},
\end{equation}
where the estimation error approaches zero as $N\to\infty$ \cite{neal2001annealed}.

The main difficulty in above computations lies in the intractability of the marginal \textcolor{black}{encoder density} \(p_E(z_0|x)\) when calculating the weight function. To overcome this, following the conclusion from Sec.~4.3 in \citep{key-2}, we construct the following probability flow ODE:
\begin{equation}\label{eq:probability-flow}
    \mathrm{d}z_t = \left(f(z_t,t) - \frac{1}{2}g(t)^2 s(z_t,x,t;\theta)\right)\mathrm{d}t, 
\end{equation}
with the boundary condition \(z_T \sim p_E(z_T)\). 
\textcolor{black}{If \(s(z_t,x,t;\theta)\) accurately approximates the score function \(\nabla_{z_t} \log p_D(z_t|x)\) after training, the conditional distribution of $z_t|x$ given by the ODE will match that of the encoding process for each \(t \in [0,T]\). Consequently, we can employ the neural ODE method \citep{ode} to efficiently compute \(p_E(z_0|x)\) as}
\begin{eqnarray} \label{eq:ode}
\log p_E(z_0|x) = \log p_E(z_T)+\int_0^T \nabla \cdot \left(f(z_t,t) - \frac{1}{2}g(t)^2 s(z_t,x,t;\theta)\right) \mathrm{d} t.
\end{eqnarray}
\textcolor{black}{See \ref{append:ode} for details of the probability flow ODE calculation,}
\textcolor{black}{and the reweighting algorithm is summarized in Algorithm} \ref{alg:sampling}. 

\begin{algorithm}[b!]
    \caption{\textcolor{black}{Sampling and reweighting procedure}}\label{alg:sampling}
    \begin{algorithmic}
        \State \textbf{Input:} Initial density: $p_{D}(z_{0})$; Decoder: $p_{D}(\cdot|z_{0};\phi^*)$; Score: $s\left(\cdot;\theta^*\right)$; Observation: $O(\cdot)$.
        \For{$i=1,..., \text{sample\ size}$}
        \State Sample $z_0^i\sim p_{D}(z_{0}^i)$ and $x^i\sim p_{D}(x^i|z_{0}^i;\phi^*)$;
        \State Compute $O_i=O(x^i)$;
        \State Compute $p_E(z_0^i|x^i;\theta^*)$ by Eqs.~(\ref{eq:probability-flow}, \ref{eq:ode});
        \State Compute $w_i=\frac{\exp(-U(x^i))p_{E}(z_{0}^i|x^i;\theta^*)}{p_{D}(z_{0}^i)p_{D}(x^i|z_{0}^i;\phi^*)}$;
        \EndFor
        \State Estimate $\mathbb E_{\pi}[\hat{O}(x)]=\frac{\sum_i w_i O_i}{\sum_i w_i}$;
        \State \textbf{Output:} $\{x_i\}$, $\mathbb E_{\pi}[\hat{O}(x)]$ 
    \end{algorithmic}
\end{algorithm}

In addition, the weight function \(w\) can also be used to estimate the normalizing constant \(Z\), which is a crucial task in many applications, such as Bayesian model selection in statistics and free energy estimation in statistical physics. Based on Eq.~\eqref{eq:vae-loss}, we have:
\begin{equation}\label{eq:variational-logZ}
\log Z \geq \mathbb{E}_{p_{D}(x,z_0)}\left[\log w(x, z_0)\right].
\end{equation}
\textcolor{black}{Here, the lower bound can also be estimated using samples from the decoder, and the tightness of this bound is achieved when \(p_{D}(x, z_0) = p_{E}(x, z_0)\). In practice, the estimate of $\log Z$ given by the above inequality can serve as an indicator of training quality. A larger value suggests a more accurate approximation and, consequently, better model performance. More detailed analysis is provided in \ref{append:logZ}.}

\subsection{Network design}\label{sec:network}
Below, we present the construction details of modules in EDG used in our experiments. 
In practical applications, more effective neural networks can be designed as needed.
\subsubsection{Boundary condition-guided score function model}
Considering that the true score function satisfies the following boundary conditions for $t=0$ and $T$:
\begin{eqnarray*}
\nabla_{z_{0}}\log p_{D}(z_{0}|x) & = & \nabla_{z_{0}}\left[\log p_{D}(z_{0}|x)+\log p_{D}(x)\right]\\
 & = & \nabla_{z_{0}}\log p_{D}(x,z_{0})\\
 & = & \nabla_{z_{0}}\left[\log p_{D}(x|z_{0})+\log p_{D}(z_{0})\right]
\end{eqnarray*}
and
\[
\nabla_{z_{T}}\log p_{D}(z_{T}|x)=\nabla_{z_{T}}\log p_{D}(z_{T}),
\]
we propose to express $s(z,x,t;\theta)$ as
\begin{eqnarray*}
s(z,x,t;\theta) & = & \left(1-\frac{t}{T}\right)\cdot\nabla_{z_{0}}\left[\log p_{D}(x|z_{0}=z)+\log p_{D}(z_{0}=z)\right] \\
 & & +\frac{t}{T}\cdot\nabla_{z_{T}}\log p_{D}(z_{T}=z)+\frac{t}{T}\left(1-\frac{t}{T}\right)s^{\prime}(z,x,t;\theta),
\end{eqnarray*}
where $s^{\prime}(z,x,t;\theta)$ is the neural network to be trained.
This formulation ensures that the error of $s$ is zero for both $t=0$ and $t=T$.

\subsubsection{Generalized Hamiltonian dynamics-based decoder}\label{sec:GHD}
Inspired by generalized Hamiltonian dynamics (GHD) \citep{mc-4,mc-5}, the decoder generates the output \(x\) using the following process. First,  an initial sample and velocity (\(y, v\)) are generated according to the latent variable \(z_0\). Then, \((y,v)\) is iteratively updated as follows:
\begin{eqnarray*}
v & := & v - \frac{\epsilon_0e^{\epsilon_0\epsilon(l;\phi)}}{2} \left( \nabla U(y) \odot e^{\frac{\epsilon_0}{2} Q_v(y, \nabla U(y), l; \phi)} + T_v(y, \nabla U(y), l; \phi) \right), \\
y & := & y + \epsilon_0 e^{\epsilon_0\epsilon(l;\phi)} \left( v_k \odot e^{\epsilon_0 Q_y(v_k, l; \phi)} + T_y(v_k, l; \phi) \right), \\
v & := & v - \frac{\epsilon_0e^{\epsilon_0\epsilon(l;\phi)}}{2} \left( \nabla U(y) \odot e^{\frac{\epsilon_0}{2} Q_v(y, \nabla U(y), l; \phi)} + T_v(y, \nabla U(y), l; \phi) \right).
\end{eqnarray*}
Finally, the decoder output \(x\) is given by:
\[
x = y - \epsilon_0 e^{\epsilon_0\eta(y; \phi)} \nabla U(y) + \sqrt{2 \epsilon_0e^{\epsilon_0\eta(y; \phi)}} \xi,
\]
where \(\xi\) is distributed according to the standard Gaussian distribution. The equation can be interpreted as a finite-step approximation of Brownian dynamics \( \mathrm{d}y = -\nabla U(y) \, \mathrm{d}t + \mathrm{d}W_t\). 
\textcolor{black}{The decoder density is then defined as}
\[
p_{D}(x | z_{0}; \phi) = \mathcal{N}\left(x \mid y - \epsilon_0e^{\epsilon_{0} \eta(y; \phi)} \nabla U(y), \, 2 \epsilon_0 e^{\epsilon_{0} \eta(y; \phi)} I\right),
\]
\textcolor{black}{where \(I\) denotes the identity matrix. Since this density can be tractably evaluated given \(x\), \(z_0\), and \(\phi\), all decoder parameters \(\phi\) involved in the GHD module can be jointly optimized with the encoder parameters \(\theta\) by minimizing the loss function \(\mathcal{L}(\theta, \phi)\).}
\textcolor{black}{In the above procedure, \(Q_v\), \(T_v\), $Q_y$, $T_y$ are all trainable neural networks. To mitigate the impact of step sizes on model performance, we parameterize them as trainable positive functions of the form \(\epsilon_0\exp(\epsilon_0 \epsilon(l; \phi))\) and \(\epsilon_0\exp(\epsilon_0 \eta(y; \phi))\), where \(\epsilon_0\) is a small positive constant that controls the initial scale and prevents instability during training caused by large step sizes. The choice of \(\epsilon_0\) in practice is discussed in \ref{append:ablation}.}

The key advantages of the GHD-based decoder are twofold. First, it effectively leverages the gradient information of the energy function, and our experiments show that it can enhance the sampling performance for multi-modal distributions. Second, by incorporating trainable correction terms and steps into the classical Hamiltonian dynamics, it achieves a good decoder density with only a few iterations. The complete configurations can be found in \ref{append:hm-decoder}.

\section{Experiments}\label{sec:experiments}

We conduct an empirical evaluation of EDG across a diverse range of energy functions. We begin with experiments on a set of two-dimensional distributions, followed by results on Bayesian logistic regression. We then assess EDG's performance on two Lennard-Jones tasks, and finally apply it to the Ising model.
All the experimental details are provided in \ref{append:experimental-details}.
Additionally, we perform an ablation study to verify the effectiveness of each module in EDG. Please refer to \ref{append:ablation} for more information.


In order to demonstrate the superiority of our model, we compare EDG with the following sampling methods:
\begin{itemize}
    \item Vanilla Hamiltonian Monte Carlo method \citep{mc-2}, denoted as V-HMC.
    \item L2HMC \cite{mc-5}, a GHD-based MCMC method with a trainable proposal distribution model.
    \item Boltzmann Generator (BG) \cite{key-4}, \textcolor{black}{a VI method that generates samples solely based on a prescribed energy function, without relying on any real data.} The surrogate distribution is modeled using RealNVP \citep{flow}.
    \item Neural Renormalization Group (NeuralRG) \citep{ising}, a method similar to BG and designed specifically for the Ising model. In this section, NeuralRG is only used for experiments of the Ising model.
    \item Path Integral Sampler (PIS) \cite{diff-6}, a diffusion-based sampling model via numerical simulation of an SDE.
\end{itemize}

We now present the experimental results for each sampling task in detail. 
\textcolor{black}{An analysis of the computational cost is provided in} \ref{append:computation-cost}.

\begin{table}[b!]
    \small
    \centering
    \vspace{-3mm}
    \caption{The Maximum Mean Discrepancy (MMD) between the samples generated by each generator and the reference samples. Details on the calculation of discrepancy can be found in \ref{append:experimental-details}.}\label{tab:result}
    \vspace{1.5mm}
    \begin{tabular}{ccccccc}
\toprule
& Mog2 & Mog2(i) & Mog6 & Mog9 & Ring & Ring5 \\
\midrule 
V-HMC & $\mathbf{0.01}$ & $1.56$ & $0.02$ & $0.04$ & $\mathbf{0.01}$ & $\mathbf{0.01}$ \\
L2HMC & $0.04$ & $0.94$ & $\mathbf{0.01}$ & $0.03$ & $0.02$ & $\mathbf{0.01}$ \\
BG & $1.90$ & $1.63$ & $2.64$ & $0.07$ & $0.05$ & $0.18$ \\
PIS & $\mathbf{0.01}$ & $1.66$ & $\mathbf{0.01}$ & $0.42$ & $\mathbf{0.01}$ & $0.78$ \\
\textbf{EDG} & $\mathbf{0.01}$ & $\mathbf{0.50}$ & $\mathbf{0.01}$ & $\mathbf{0.02}$ & $\mathbf{0.01}$ & $0.02$ \\
\bottomrule 
    & & & & &\\
    \end{tabular}
\vspace{-3mm}
\end{table}

\begin{figure*}[t!]
	\raggedright
    \vspace{-3mm}
    \subfigure{
    {\scriptsize{~~~~~~~~~\textbf{MoG2}~~~~~~~~~~~}}
    {\scriptsize{\textbf{MoG2(i)}~~~~~~~~~}}
    {\scriptsize{\textbf{MoG6}~~~~~~~~~~~}}
    {\scriptsize{\textbf{MoG9}~~~~~~~~~~~~}}
    {\scriptsize{\textbf{Ring}}~~~~~~~~~~~~}
    {\scriptsize{\textbf{Ring5}~~~~~~~~~~~}}
    }
    
        \vspace{-3mm}
	\subfigure{\rotatebox{90}{\tiny{~~~~~}}
        \rotatebox{90}{\scriptsize{~~~~~~\textbf{Ref}}}
		\begin{minipage}[t]{0.15\linewidth}
			\raggedright
			\includegraphics[width=1\linewidth]{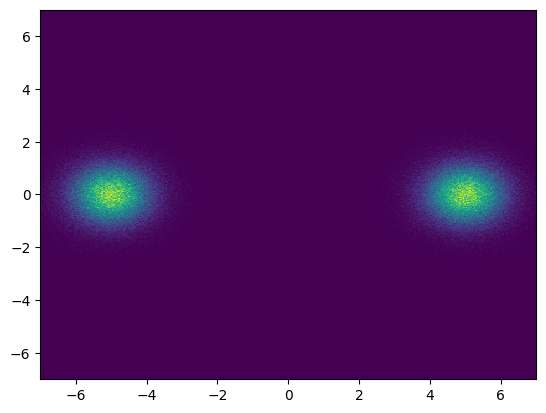}
		\end{minipage}
		\begin{minipage}[t]{0.15\linewidth}
			\raggedright
			\includegraphics[width=1\linewidth]{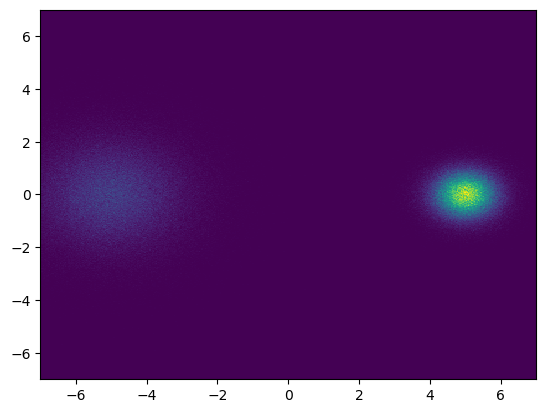}
		\end{minipage}
		\begin{minipage}[t]{0.15\linewidth}
			\raggedright
			\includegraphics[width=1\linewidth]{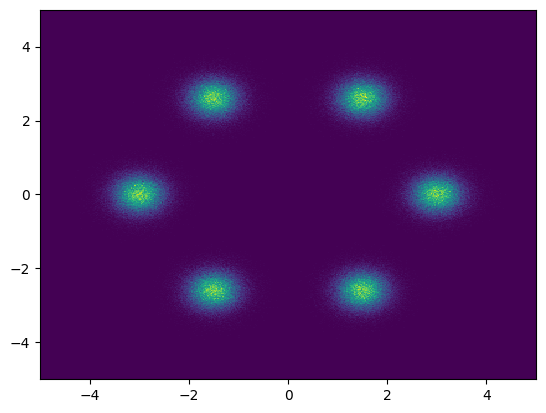}
		\end{minipage}
		\begin{minipage}[t]{0.15\linewidth}
			\raggedright
			\includegraphics[width=1\linewidth]{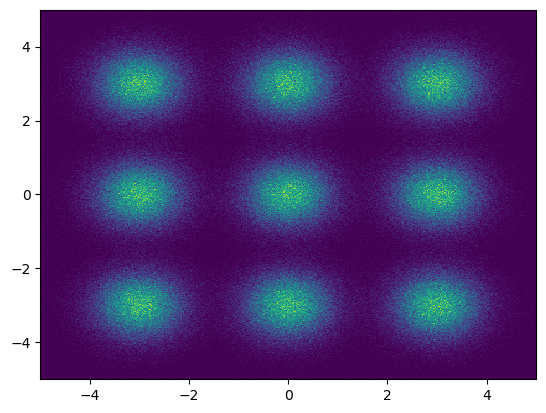}
		\end{minipage}
		\begin{minipage}[t]{0.15\linewidth}
			\raggedright
			\includegraphics[width=1\linewidth]{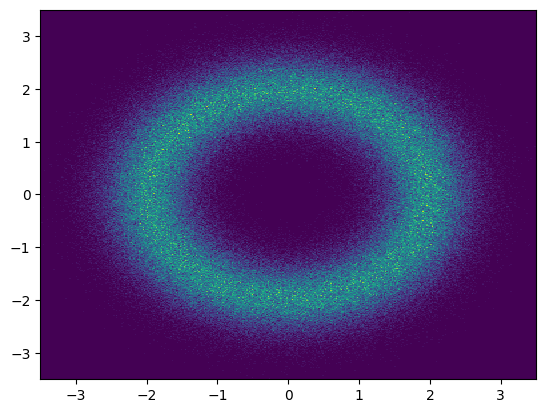}
		\end{minipage}
		\begin{minipage}[t]{0.15\linewidth}
			\raggedright
			\includegraphics[width=1\linewidth]{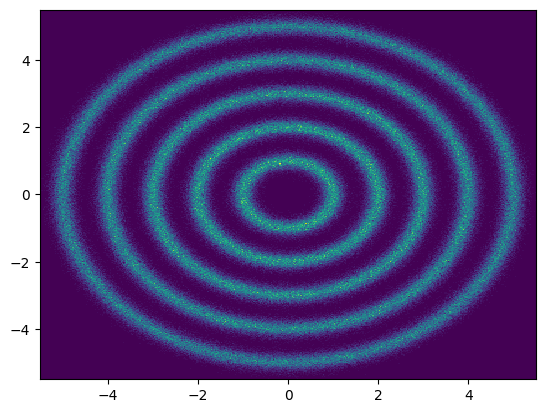}
		\end{minipage}
	}

        \vspace{-3mm}
    \subfigure{\rotatebox{90}{~~~~~}
        \rotatebox{90}{\scriptsize{~~\textbf{V-HMC}}}
		\begin{minipage}[t]{0.15\linewidth}
			\raggedright
			\includegraphics[width=1\linewidth]{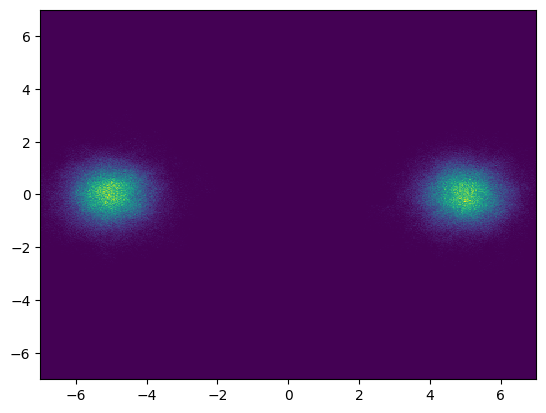}
		\end{minipage}
		\begin{minipage}[t]{0.15\linewidth}
			\raggedright
			\includegraphics[width=1\linewidth]{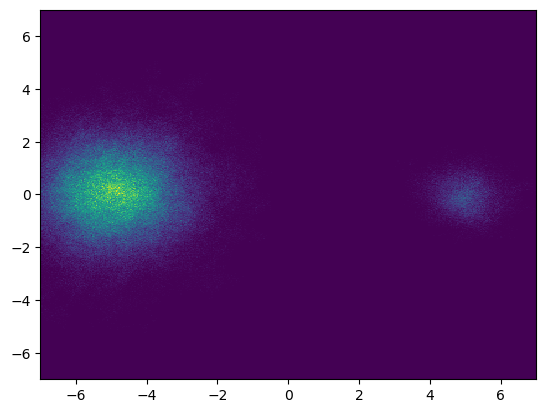}
		\end{minipage}
		\begin{minipage}[t]{0.15\linewidth}
			\raggedright
			\includegraphics[width=1\linewidth]{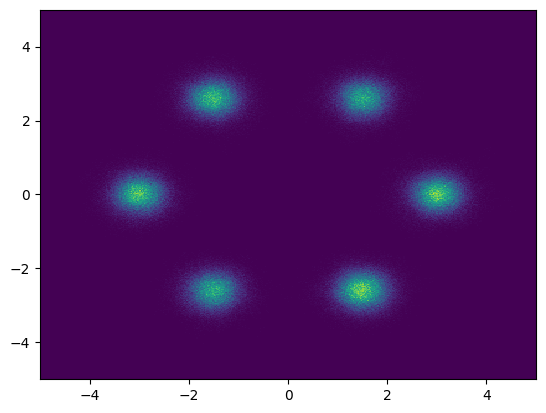}
		\end{minipage}
		\begin{minipage}[t]{0.15\linewidth}
			\raggedright
			\includegraphics[width=1\linewidth]{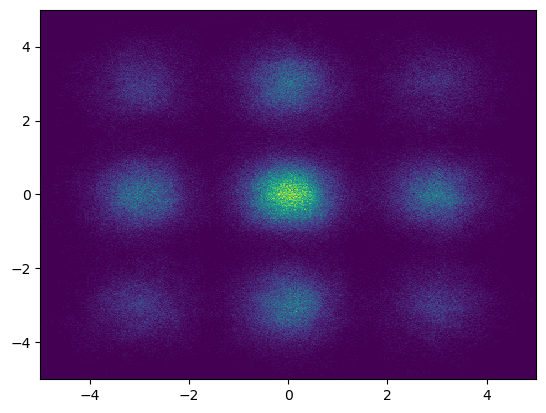}
		\end{minipage}
		\begin{minipage}[t]{0.15\linewidth}
			\raggedright
			\includegraphics[width=1\linewidth]{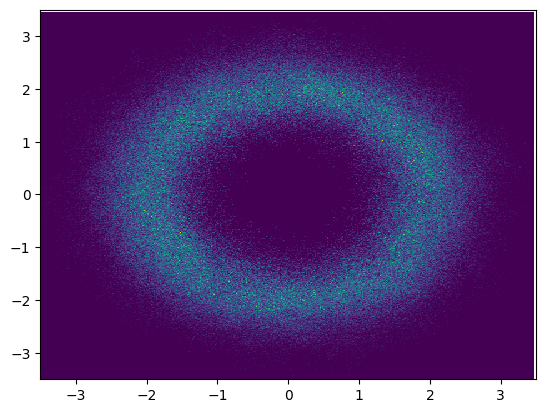}
		\end{minipage}
		\begin{minipage}[t]{0.15\linewidth}
			\raggedright
			\includegraphics[width=1\linewidth]{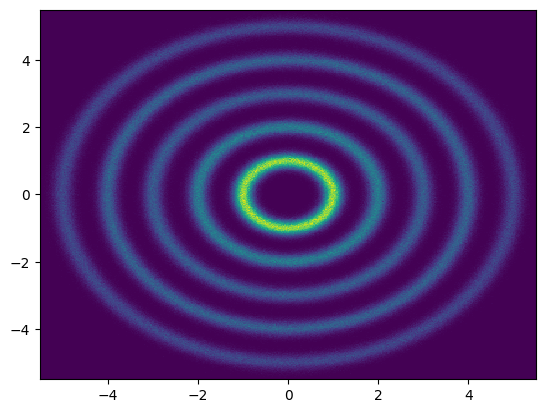}
		\end{minipage}
	}

        \vspace{-3mm}
    \subfigure{\rotatebox{90}{~~~~~}
        \rotatebox{90}{\scriptsize{~~\textbf{L2HMC}}}
		\begin{minipage}[t]{0.15\linewidth}
			\raggedright
			\includegraphics[width=1\linewidth]{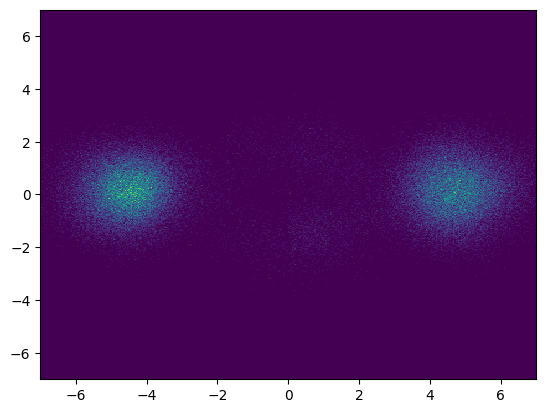}
		\end{minipage}
		\begin{minipage}[t]{0.15\linewidth}
			\raggedright
			\includegraphics[width=1\linewidth]{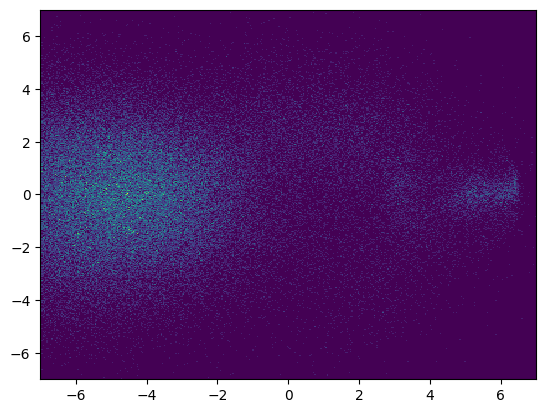}
		\end{minipage}
		\begin{minipage}[t]{0.15\linewidth}
			\raggedright
			\includegraphics[width=1\linewidth]{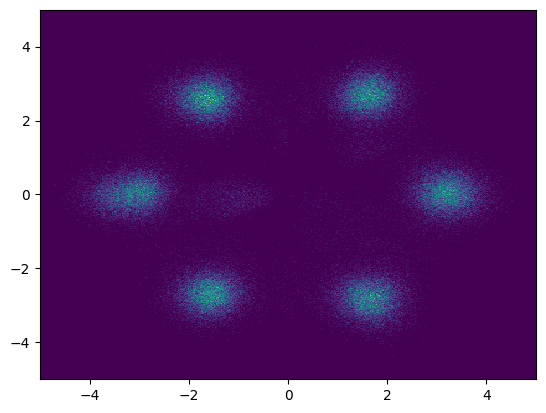}
		\end{minipage}
		\begin{minipage}[t]{0.15\linewidth}
			\raggedright
			\includegraphics[width=1\linewidth]{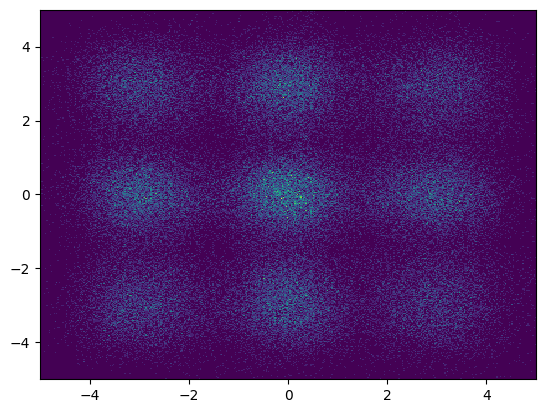}
		\end{minipage}
		\begin{minipage}[t]{0.15\linewidth}
			\raggedright
			\includegraphics[width=1\linewidth]{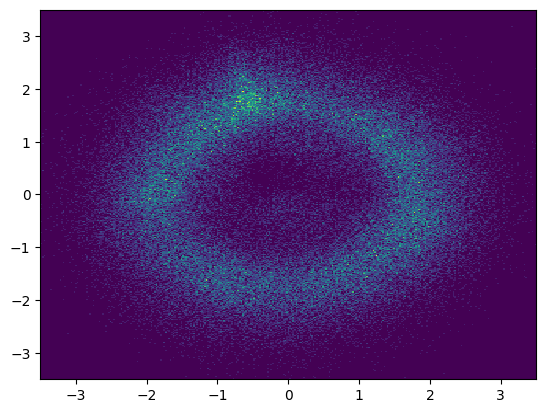}
		\end{minipage}
		\begin{minipage}[t]{0.15\linewidth}
			\raggedright
			\includegraphics[width=1\linewidth]{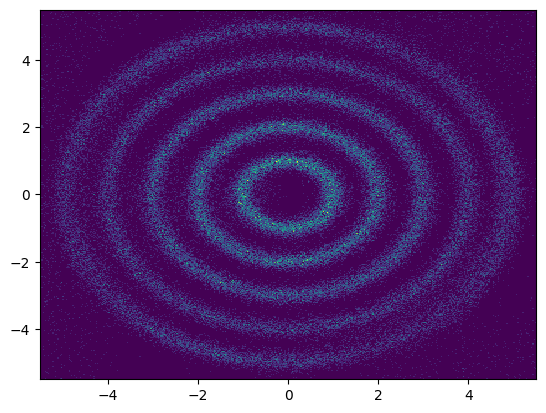}
		\end{minipage}
	}

        \vspace{-3mm}
	\subfigure{\rotatebox{90}{\tiny{~~~~~}}
        \rotatebox{90}{\scriptsize{~~~~~~~\textbf{BG}}}
		\begin{minipage}[t]{0.15\linewidth}
			\raggedright
			\includegraphics[width=1\linewidth]{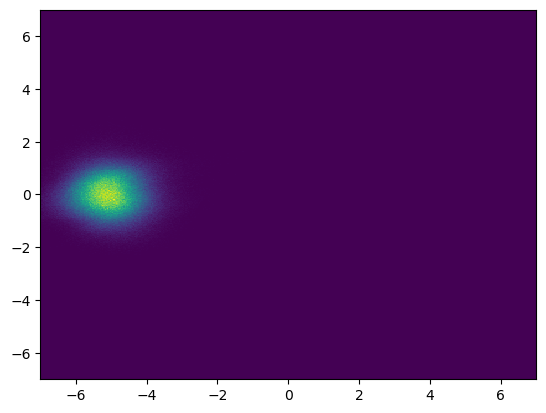}
		\end{minipage}
		\begin{minipage}[t]{0.15\linewidth}
			\raggedright
			\includegraphics[width=1\linewidth]{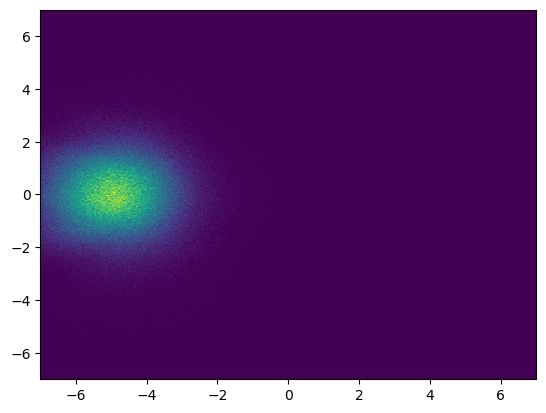}
		\end{minipage}
		\begin{minipage}[t]{0.15\linewidth}
			\raggedright
			\includegraphics[width=1\linewidth]{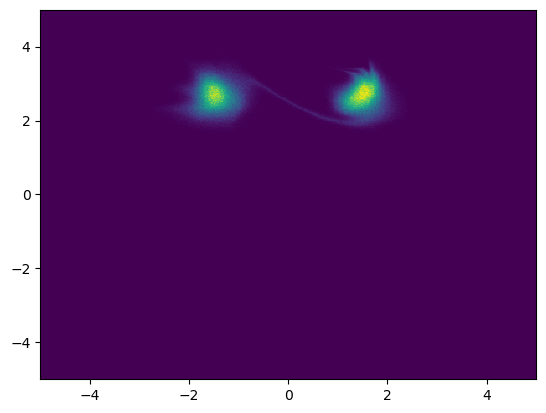}
		\end{minipage}
		\begin{minipage}[t]{0.15\linewidth}
			\raggedright
			\includegraphics[width=1\linewidth]{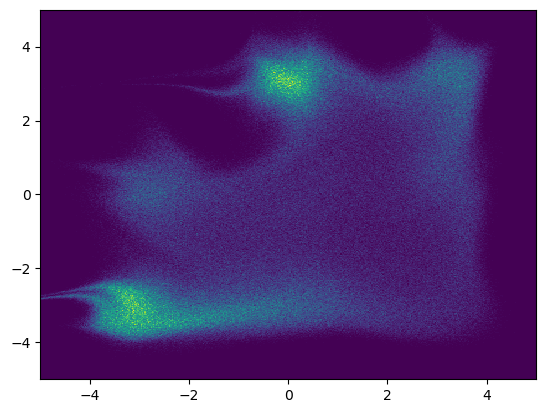}
		\end{minipage}
		\begin{minipage}[t]{0.15\linewidth}
			\raggedright
			\includegraphics[width=1\linewidth]{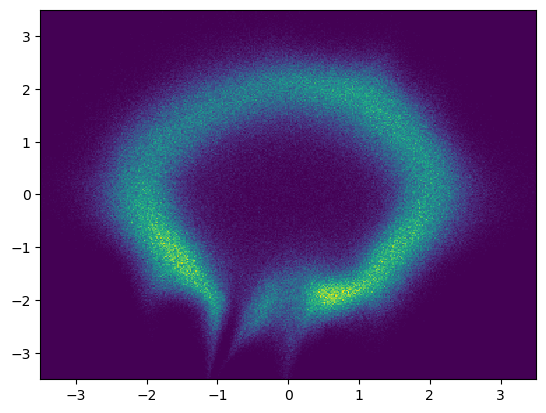}
		\end{minipage}
		\begin{minipage}[t]{0.15\linewidth}
			\raggedright
			\includegraphics[width=1\linewidth]{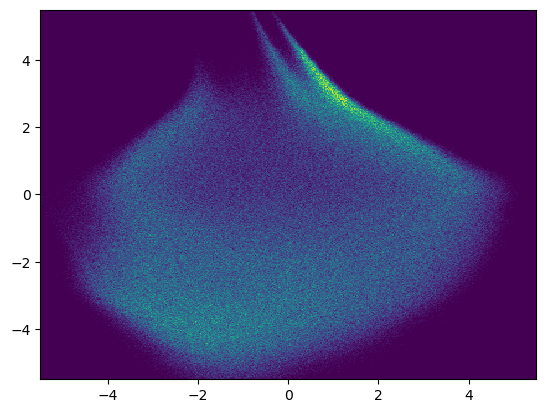}
		\end{minipage}
	}
 
        \vspace{-3mm}
	\subfigure{\rotatebox{90}{\tiny{~~~~}}
        \rotatebox{90}{\scriptsize{~~~~~~\textbf{PIS}}}
		\begin{minipage}[t]{0.15\linewidth}
			\raggedright
			\includegraphics[width=1\linewidth]{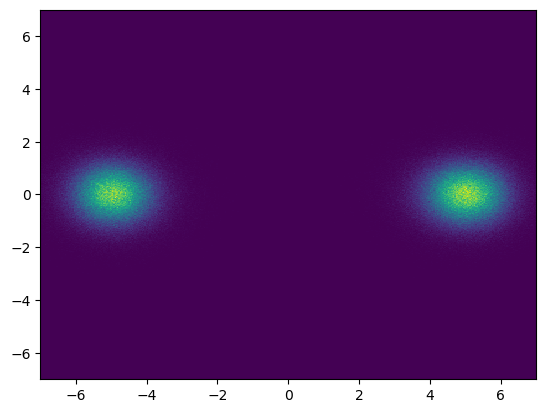}
		\end{minipage}
		\begin{minipage}[t]{0.15\linewidth}
			\raggedright
			\includegraphics[width=1\linewidth]{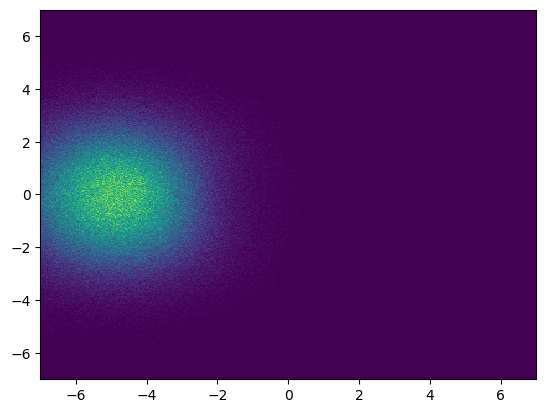}
		\end{minipage}
		\begin{minipage}[t]{0.15\linewidth}
			\raggedright
			\includegraphics[width=1\linewidth]{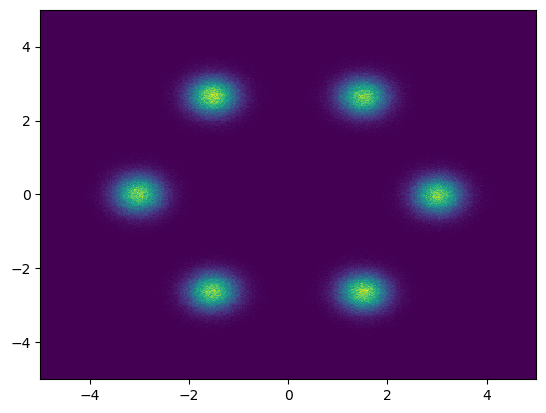}
		\end{minipage}
		\begin{minipage}[t]{0.15\linewidth}
			\raggedright
			\includegraphics[width=1\linewidth]{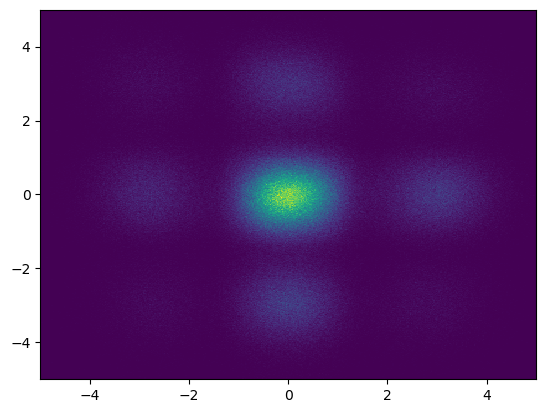}
		\end{minipage}
		\begin{minipage}[t]{0.15\linewidth}
			\raggedright
			\includegraphics[width=1\linewidth]{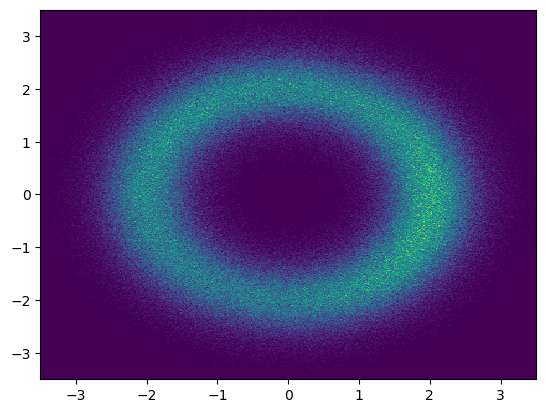}
		\end{minipage}
		\begin{minipage}[t]{0.15\linewidth}
			\raggedright
			\includegraphics[width=1\linewidth]{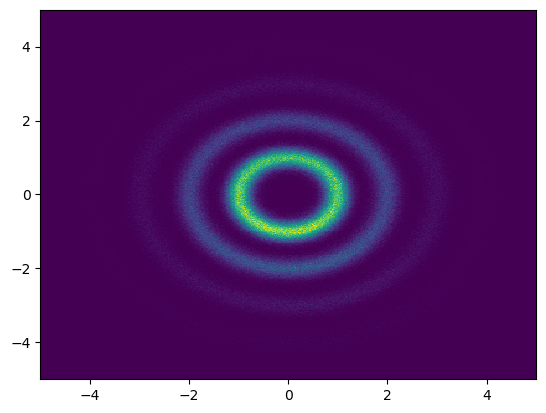}
		\end{minipage}
	}
 
 
	   \vspace{-3mm}
        \subfigure{\rotatebox{90}{\tiny{~~~~}}
        \rotatebox{90}{\scriptsize{~~~~~\textbf{EDG}}}
		\begin{minipage}[t]{0.15\linewidth}
			\raggedright
			\includegraphics[width=1\linewidth]{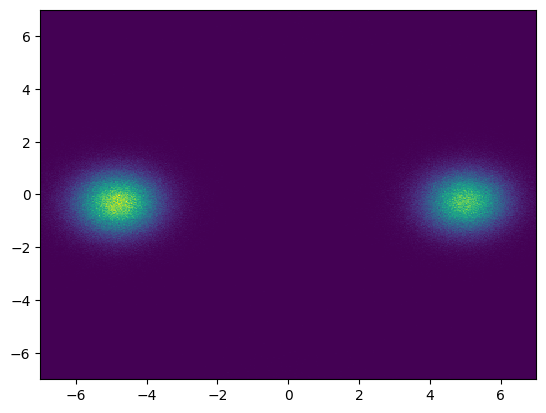}
		\end{minipage}
		\begin{minipage}[t]{0.15\linewidth}
			\raggedright
			\includegraphics[width=1\linewidth]{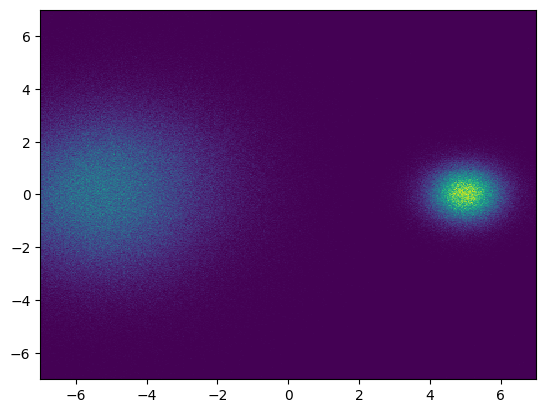}
		\end{minipage}
		\begin{minipage}[t]{0.15\linewidth}
			\raggedright
			\includegraphics[width=1\linewidth]{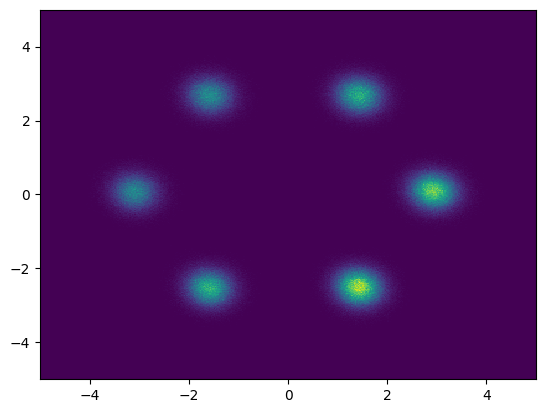}
		\end{minipage}
		\begin{minipage}[t]{0.15\linewidth}
			\raggedright
			\includegraphics[width=1\linewidth]{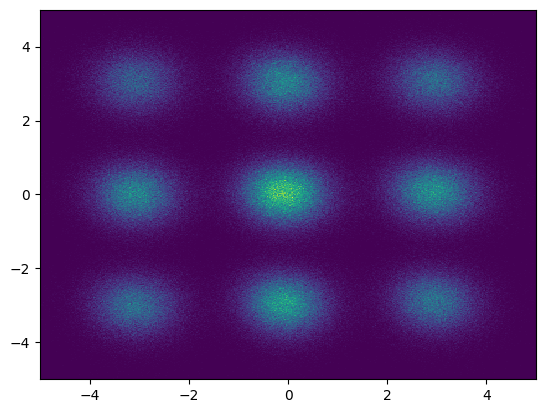}
		\end{minipage}
		\begin{minipage}[t]{0.15\linewidth}
			\raggedright
			\includegraphics[width=1\linewidth]{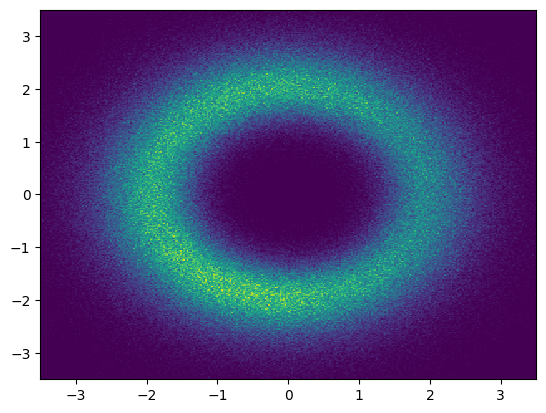}
		\end{minipage}
		\begin{minipage}[t]{0.15\linewidth}
			\raggedright
			\includegraphics[width=1\linewidth]{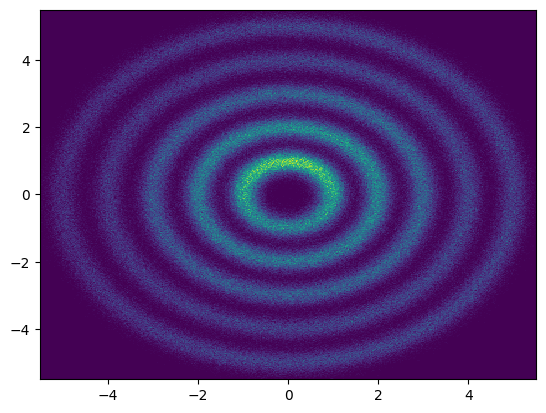}
		\end{minipage}
	}
    \vspace{-8mm}
	\caption{Density plots for 2D energy function. For the generation of reference samples, please refer to \ref{append:experimental-details}. We generate $500,000$ samples for each method and plot the histogram.}
	\label{fig:2d-example}
\end{figure*}

\textbf{2D energy function} Firstly, we compare our model with the other models on several synthetic 2D energy functions: MoG2 (Mixture of two isotropic Gaussians with equal variance $\sigma^2=0.5$, centers $10$ apart), MoG2(i) (Mixture of two isotropic Gaussians with unequal variance $\sigma_1^2=1.5, \sigma_2^2=0.3$, centers $10$ apart), MoG6 (Mixture of six isotropic Gaussians with variance $\sigma^2=0.1$), MoG9 (Mixture of nine isotropic Gaussians with variance $\sigma^2=0.3$), Ring, Ring5 (energy functions can be seen in \cite{mc-4}). 
We present the histogram of samples for visual inspection in Fig.~\ref{fig:2d-example}, and Table \ref{tab:result} summarizes the sampling errors.
As shown, EDG delivers higher-quality samples compared to the other methods. 
To further clarify the contribution of each component in EDG, we compare it with a vanilla VAE in \ref{append:ablation}. The effectiveness of the reweighting scheme is also evaluated in \ref{append:experimental-details}.

\textbf{Bayesian Logistic Regression } 
In the subsequent experiments, we highlight the efficacy of EDG in the context of Bayesian logistic regression, particularly when dealing with a posterior distribution residing in a higher-dimensional space. In this scenario, we tackle a binary classification problem with labels $L=\{0, 1\}$ and high-dimensional features $D$. The classifier's output is defined as
\[
p(L=1|D,x)=\mathrm{softmax}(w^\top D+b),
\]
where $x=(w,b)$. We aim to draw samples $x^1,\ldots,x^N$ from the posterior distribution
\[
\pi(x)\propto p(x)\prod_{(L,D)\in\mathcal D_{\mathrm{train}}}p(L|D,x)
\]
based on the training set $\mathcal D_{\mathrm{train}}$, where the prior distribution $p(x)$ is a standard Gaussian distribution. Then, for a given $D$, the conditional distribution $p(L|D,\mathcal{D}_{\mathrm{train}})$ can be approximated as $\frac{1}{N}\sum_{n}p(L|D,x^{n})$. We conduct experiments on three datasets: Australian (AU, 15 covariates), German (GE, 25 covariates), and Heart (HE, 14 covariates) \citep{logistic1}, evaluating accuracy rate (ACC) and Area Under the Curve (AUC) on the test subset. Remarkably, as illustrated in Table \ref{table:bayes}, EDG consistently achieves the highest accuracy and AUC performance.

\begin{table}[t!]
\vspace{-3mm}
\begin{centering}
\caption{Classification accuracy and AUC results for Bayesian logistic regression
tasks. The experiments utilize consistent training and test data partitions,
where the HMC step size is set to $0.01$. Average accuracy and AUC values,
accompanied by their respective standard deviations, are computed
across $32$ independent experiments for all datasets.\label{table:bayes}}
\vspace{1.5mm}
\resizebox{\textwidth}{!}
{
\begin{tabular}{ccccccc}
\toprule 
\multirow{2}{*}{} & \multicolumn{2}{c}{{\footnotesize{}AU}} & \multicolumn{2}{c}{{\footnotesize{}GE}} & \multicolumn{2}{c}{{\footnotesize{}HE}}\tabularnewline
\cmidrule{2-7} \cmidrule{3-7} \cmidrule{4-7} \cmidrule{5-7} \cmidrule{6-7} \cmidrule{7-7} 
 & {\footnotesize{}Acc} & {\footnotesize{}Auc} & {\footnotesize{}Acc} & {\footnotesize{}Auc} & {\footnotesize{}Acc} & {\footnotesize{}Auc}\tabularnewline
\midrule 
{\scriptsize{}V-HMC} & {\scriptsize{}$82.97\pm1.94$} & {\scriptsize{}$90.88\pm0.83$} & {\scriptsize{}$78.52\pm0.48$} & {\scriptsize{}$77.67\pm0.28$} & {\scriptsize{}$86.75\pm1.63$} & {\scriptsize{}$93.35\pm0.76$}\tabularnewline
{\scriptsize{}L2HMC} & {\scriptsize{}$73.26\pm1.56$} & {\scriptsize{}$79.69\pm3.65$} & {\scriptsize{}$62.02\pm4.19$} & {\scriptsize{}$60.23\pm5.10$} & {\scriptsize{}$82.23\pm2.81$} & {\scriptsize{}$90.48\pm0.51$}\tabularnewline
{\scriptsize{}BG} & {\scriptsize{}$82.99\pm1.18$} & {\scriptsize{}$91.23\pm0.67$} & {\scriptsize{}$78.14\pm1.44$} & {\scriptsize{}$77.59\pm0.73$} & {\scriptsize{}$86.75\pm1.99$} & {\scriptsize{}$93.44\pm0.39$}\tabularnewline
{\scriptsize{}PIS} & {\scriptsize{}$81.64\pm2.63$} & {\scriptsize{}$91.23\pm0.67$} & {\scriptsize{}$71.90\pm3.17$} & {\scriptsize{}$71.67\pm4.52$} & {\scriptsize{}$83.24\pm3.95$} & {\scriptsize{}$91.68\pm2.78$}\tabularnewline
{\scriptsize{}\textbf{EDG}} & {\scriptsize{}$\mathbf{84.96\pm1.67}$} & {\scriptsize{}$\mathbf{92.82\pm0.69}$} & {\scriptsize{}$\mathbf{79.40\pm1.74}$} & {\scriptsize{}$\mathbf{82.79\pm1.46}$} & {\scriptsize{}$\mathbf{88.02\pm3.90}$} & {\scriptsize{}$\mathbf{95.10\pm1.23}$}\tabularnewline
\bottomrule
\end{tabular}
}
\end{centering}
\end{table}

\begin{table}[t!]
\scriptsize
\vspace{-3mm}
\begin{centering}
\caption{Classification accuracy on the test dataset of Covertype. The reported values represent averages and standard deviations of accuracy over 32 independent experiments.\label{table:bayes2}}
\vspace{1.5mm}
\begin{tabular}{ccccccc}
\toprule 
 & V-HMC & L2HMC & BG & PIS & \textbf{EDG} \tabularnewline
\midrule 
Acc & $49.88\pm3.32$ & $51.51\pm3.46$ & $50.75\pm 3.78$ & $50.59\pm2.94$ & $\mathbf{70.13\pm2.13}$\tabularnewline
\bottomrule
\end{tabular}
\par\end{centering}
\end{table}

We extend our analysis to the binary Covertype dataset comprising 581,012 data points and 54 features. The posterior of the classifier parameters follows a hierarchical Bayesian model (see Sec.~5 of \citep{vi-7}), with $x$ denoting the combination of classifier parameters and the hyperparameter in the hierarchical Bayesian model. To enhance computational efficiency, in BG and EDG, $\log\pi(x)$ is unbiasedly approximated during training as
\[
\log\pi(x)\approx \log p(x)+\frac{|\mathcal D_{\mathrm{train}}|}{|\mathcal B|}\sum_{(L,D)\in\mathcal B}\log p(L|D,x),
\]
where $\mathcal B$ is a random mini-batch. For V-HMC and L2HMC, the exact posterior density is calculated. As indicated by the results in Table \ref{table:bayes2}, EDG consistently outperforms alternative methods.

\textbf{Lennard-Jones } We further evaluate the performance of EDG on two \textnormal{Lennard--Jones} (LJ) systems \textcolor{black}{with non-periodic boundary conditions}, consisting of 13 and 55 particles (LJ13 and LJ55), as described in \cite{klein2023equivariant}. The LJ potential is a commonly used interaction model that captures both repulsive and attractive forces between non-bonded particles. The potential energy depends on the pairwise distances between particles, and its explicit form is provided in \ref{append:experimental-details}. In this task, we temporarily disregard the issue of model equivariance, which we leave for future investigation.

\begin{figure}[h!]
\begin{center}
\includegraphics[width=\textwidth]{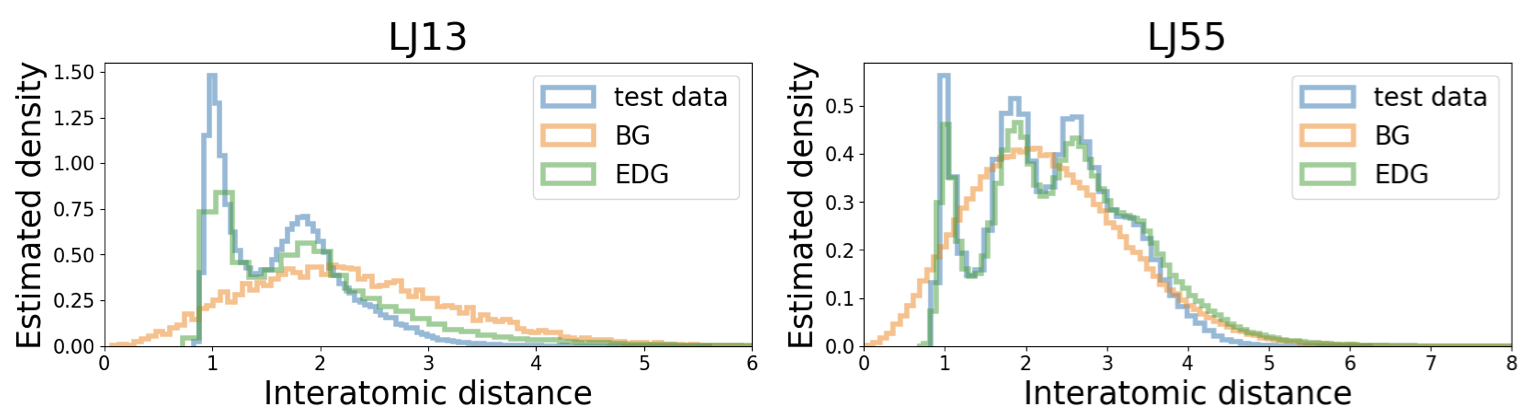}
\end{center}
\vspace{-7mm}
\caption{Comparison of the test data interatomic distance of LJ13 (left), LJ55 (right) with samples generated from BG and EDG. The results for L2HMC and PIS are provided separately in \ref{append:experimental-details} due to significant deviations from the test data.}
\label{fig:LJ-results}
\end{figure}

LJ13 refers to the system of 13 particles, $x= \{ x_1,...,x_{13}\}$ with 3 dimensions each, resulting in a task with dimensionality $d = 39$. LJ55 meanwhile refers to a system of 55 particles, $x = \{x_1,..., x_{55}\}$ with 3 dimensions each, resulting in a high dimensional task with dimensionality $d = 165$. For the experimental results, we evaluate the generated sample with the test data from \cite{klein2023equivariant} obtained by MCMC. 
A visualization of the interatomic distances is presented in Fig.~\ref{fig:LJ-results}. In \ref{append:experimental-details}, we test the effectiveness of sample reweighting introduced in Sec. \ref{sec:marginal-encoder-density}, by using the \textcolor{black}{relative Effective Sample Size (see Eq.~\eqref{eq:ress} and Table \ref{table:ess}).}

\begin{figure}[b!]
	\begin{center}
\includegraphics[width=1.\textwidth]{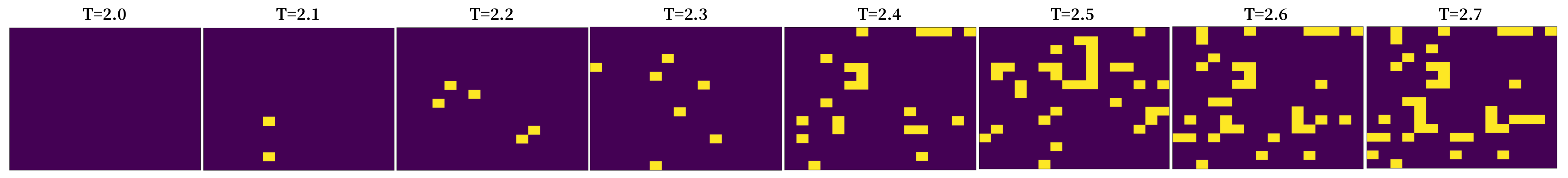}
\end{center}
\vspace{-7mm}
    \caption{The generated states at different temperatures from $T=2.0$ to $T=2.7$ by EDG with a dimensionality of 256 ($16\times 16$), where the latent variable $z_0$ remains fixed. As the temperature rises, the model's state tends progressively toward disorder.}
	\label{fig:ising}
\end{figure}

\begin{table}[t!]
\small
\begin{centering}
\caption{The estimation of $\log Z_{\mathrm{Ising}}$ in 2D ising model with a dimension of 256 ($16\times 16$) is obtained by the method described in Sec.~\ref{sec:marginal-encoder-density}. We utilize a batch size of $n=256$ to estimate the mean and applies the central limit theorem to calculate the standard deviation of the mean of the statistic as $\mathrm{std}/\sqrt{n}$.}\label{table:ising}
\vspace{1.5mm}
\begin{tabular}{ccccc}
\toprule 
$\log Z_{\mathrm{Ising}}$ & NeuralRG & PIS & \textbf{EDG} 
\tabularnewline
\midrule 
$T=2.0$ & $260.24\pm0.13$ & $210.17\pm0.43$ & $\mathbf{270.32\pm0.18}$ \\ 
$T=2.1$ & $250.57\pm0.14$ & $208.48\pm0.41$ & $\mathbf{260.34\pm0.19}$ \\ 
$T=2.2$ & $239.43\pm0.16$ & $210.57\pm0.39$ & $\mathbf{252.11\pm0.17}$ \\ 
$T=2.3$ & $231.25\pm0.15$ & $214.64\pm0.37$ & $\mathbf{233.46\pm0.17}$ \\ 
$T=2.4$ & $\mathbf{225.69\pm0.17}$ & $212.43\pm0.37$ & $225.02\pm0.15$ \\ 
$T=2.5$ & $219.26\pm0.17$ & $202.92\pm0.37$ & $\mathbf{220.03\pm0.14}$ \\ 
$T=2.6$ & $\mathbf{216.69\pm0.18}$ & $181.83\pm0.40$ & $214.78\pm0.14$ \\ 
$T=2.7$ & $212.42\pm0.18$ & $189.16\pm0.36$ & $\mathbf{212.57\pm0.14}$ 
\tabularnewline
\bottomrule
\end{tabular}
\par\end{centering}
\end{table}

\textbf{Ising model } Finally, we verify the performance of EDG on the 2-dimensional Ising model \cite{ising}, a mathematical model of ferromagnetism in statistical mechanics. To ensure the continuity of physical variables, we employ a continuous relaxations trick \citep{isingrelax} to transform discrete variables into continuous auxiliary variables with the target distribution:
\[
\pi(x)=\exp \left(-\frac{1}{2} x^T\left(K(T)+\alpha I\right)^{-1} x\right) \times \prod_{i=1}^N \cosh \left(x_i\right),
\]
where $K$ is an $N\times N$ symmetric matrix depending on the temperature $T$, $\alpha$ is a constant guaranteeing $K+\alpha I$ to be positive. 
For the corresponding discrete Ising variables $\boldsymbol{s}=\{1,-1\}^{\otimes N}$, one can directly obtain discrete samples according to $\pi(\boldsymbol{s}|\boldsymbol{x})=\prod_i\left(1+e^{-2 s_i x_i}\right)^{-1}$.
When there is no external magnetic field and each spin can only interact with its neighboring spins, $K$ is defined as $\sum_{<ij>}s_is_j / T$, with the nears neighboring sum $<ij>$.
Consequently, the normalizing constant of the continuous relaxations system is given by $\log Z=\log Z_{\text {Ising }} + \frac{1}{2} \ln \operatorname{det}(K+\alpha I) - \frac{N}{2}[\ln (2 / \pi)-\alpha]$~\cite{ising}.
Additionally, using the method described in Sec.~\ref{sec:marginal-encoder-density}, we provide the estimates of $\log Z_{\text{Ising}}$ at different temperatures for samples generated by NeuralRG, PIS and EDG. Since these are lower bound estimates, larger values indicate more accurate results. As seen in Table \ref{table:ising}, EDG provides the most accurate estimates of $\log Z$ across most temperature ranges. The generated states at different temperatures are displayed in Fig. \ref{fig:ising}.

\section{Conclusion}\label{sec:conclusion}
\vspace{-1mm}
Our work introduces the EDG as an innovative and effective sampling approach by combining principles from VI and diffusion based methods. Drawing inspiration from VAEs, EDG excels in efficiently generating samples from intricate Boltzmann distributions. Leveraging the expressive power of the diffusion model, our method accurately estimates the KL divergence without the need for numerical solutions to ordinary or stochastic differential equations. Empirical experiments validate the superior sampling performance of EDG.

In consideration of its powerful generation ability and unrestrained network design theoretically, there is still room for further exploration. We can design specific network architectures for different tasks.
\textcolor{black}{Regarding many-body particle systems, we plan to leverage equivariant graph neural networks (EGNN) \cite{kohler2020equivariant,klein2023equivariant} to design all trainable components within the GHD module of the decoder to ensure the invariance.
Moreover, we aim to construct equivariant score function models within the encoding process with respect to both $x$ and $z_t$, which is expected to guarantee the invariance of the weight function.}
\textcolor{black}{Furthermore, theoretical convergence guarantees can be strengthened by investigating error bounds and convergence rates under finite training time and limited model capacity.}

In summary, our future work will extend the application of EDG to construct generative models for large-scale physical and chemical systems, such as proteins \citep{protein-1,protein-2}, \textcolor{black}{while simultaneously refining the theoretical foundations of our model.}

\newpage
\section*{Acknowledgements}
The first and third authors are supported by the NSF of China (under grant number 12171367). The second author is supported by the NSF of China (under grant numbers 92270115, 12071301) and the Shanghai Municipal Science and Technology Commission (No. 20JC1412500), and Henan Academy of Sciences. The last author was supported by the NSF of China (No. 12288201), and the Research Startup Fund of Henan Academy of Sciences (No. 232019024).

\bibliography{elsarticle}
\bibliographystyle{elsarticle-num-names}

\appendix
\section{Proof of Eq.~\eqref{eq:vae-loss}}\label{append:proof-vae-loss}

For completeness, here we provide a proof of Eq.~\eqref{eq:vae-loss}, which follows a similar derivation as the loss function of the standard VAE~\cite{kingma2019introduction}.  
For simplicity of notation, we omit the explicit dependence on the distribution parameters $\phi$ and $\theta$.

The KL divergence between $p_{D}(x,z_{0})$ and $p_{E}(x,z_{0})$ can be written as
\begin{eqnarray*}
D_{\mathrm{KL}}\left(p_{D}(x,z_{0})||p_{E}(x,z_{0})\right) & = & \mathbb{E}_{p_{D}}\left[\log\frac{p_{D}(x,z_{0})}{p_{E}(x,z_{0})}\right]\\
 & = & \mathbb{E}_{p_{D}}\left[\log\frac{p_{D}(x)}{\pi(x)}\right]+\mathbb{E}_{p_{D}}\left[\log\frac{p_{D}(z_{0}|x)}{p_{E}(z_{0}|x)}\right]\\
 & = & D_{\mathrm{KL}}\left(p_{D}(x)||\pi(x)\right)\\
 &  & +D_{\mathrm{KL}}\left(p_{D}(z_{0}|x)||p_{E}(z_{0}|x)\right),
\end{eqnarray*}
where
\[
D_{\mathrm{KL}}\left(p_{D}(z_{0} | x)\,\|\,p_{E}(z_{0} | x)\right) \triangleq \mathbb{E}_{p_{D}}\left[\log\frac{p_{D}(z_{0} | x)}{p_{E}(z_{0} | x)}\right]
\]
is the expected KL divergence between the conditional distributions $p_{D}(z_{0} | x)$ and $p_{E}(z_{0} | x)$ under $x \sim p_{D}(x)$, and is therefore non-negative.

Consequently, we obtain
\begin{eqnarray*}
D_{\mathrm{KL}}\left(p_{D}(x)\,\|\,\pi(x)\right)
&\leq& D_{\mathrm{KL}}\left(p_{D}(x,z_{0})\,\|\,p_{E}(x,z_{0})\right) \\
&=& D_{\mathrm{KL}}\left(p_{D}(z_{0})p_{D}(x | z_{0})\,\|\,\pi(x)p_{E}(z_{0} | x)\right),
\end{eqnarray*}
where the inequality becomes an equality if $p_{D}(z_{0} | x) = p_{E}(z_{0} | x)$ for all $x,z_0$.

\section{Analysis of the independence between $z_T$ and $x$}\label{append:independence}

In the decoding process, if $z_T$ is independent of $z_0$ with $p_D(z_{T}|z_{0})=p_D(z_T)$, we have
\begin{eqnarray*}
p_D(x,z_{T}) & = & \int p_D(x,z_{0},z_{T})\mathrm{d}z_{0}\\
 & = & \int p_D(x,z_{0})p_D(z_{T}|z_{0})\mathrm{d}z_{0}\\
 & = & p_D(z_{T})\int p_D(x,z_{0})\mathrm{d}z_{0}\\
 & = & p_D(x)p_D(z_{T}),
\end{eqnarray*}
which implies that $x$ is also independent of $z_T$.

\section{Proof of Theorem \ref{thm:upperbound}}
\label{append:upperbound}
\paragraph{Part 1}
Eq.~\eqref{eq:edg-inequality-1} can be proved in a manner similar to the proof of Eq.~\eqref{eq:vae-loss} (see \ref{append:proof-vae-loss}).
Thus, here we focus on proving that the equality in Eq.~\eqref{eq:edg-inequality-1} holds under the conditions that $z_0$ and $z_T$ are independent, and $s(z,x,t;\theta) \equiv \nabla_{z_t} \log p_{D}(z_t | x)$.

Under these assumptions, $x$ is also independent of $z_T$ in the decoding process (see \ref{append:independence}).  
Consequently, the conditional distribution of $z_{[\cdot]}$ given $x$ can be described by the reverse-time process defined by Eq.~\eqref{eq:reverse-sde}.  
Moreover, it shares the same initial distribution $p_{D}(z_T | x) = p_{D}(z_T)= p_E(z_T)$ as the reverse-time process defined by Eq.~\eqref{eq:reverse-diffusion} in the encoding process.

Furthermore, because $s(z,x,t;\theta) \equiv \nabla_{z_t} \log p_{D}(z_t | x)$, the drift and diffusion terms of the two reverse-time processes exactly match.  
Therefore, we have $p_{D}(z_{[\cdot]} | x) = p_{E}(z_{[\cdot]} | x)$, and
\begin{eqnarray*}
D_{\mathrm{KL}}\left(p_{D}(z_{[\cdot]},x)\,\|\,p_{E}(z_{[\cdot]},x)\right) 
&=& D_{\mathrm{KL}}\left(p_{D}(x)\,\|\,\pi(x)\right) \\
&& + D_{\mathrm{KL}}\left(p_{D}(z_{[\cdot]} | x)\,\|\,p_{E}(z_{[\cdot]} | x)\right) \\
&=& D_{\mathrm{KL}}\left(p_{D}(x)\,\|\,\pi(x)\right).
\end{eqnarray*}

\paragraph{Part 2}
We now prove Eq.~\eqref{eq:edg-inequality-2}.  
In the decoding process, the distribution of $z_{[\cdot]}$ can be described by the following reverse-time process:
\[
\mathrm{d}z_{\tilde{t}} = -\left(f(z_{\tilde{t}},\tilde{t}) - g(\tilde{t})^{2}\nabla_{z_{\tilde{t}}}\log p_{D}(z_{\tilde{t}})\right)\mathrm{d}\tilde{t} + g(\tilde{t})\mathrm{d}\tilde{W}_{\tilde{t}}, \ z_T\sim p_D(z_T).
\]
Combining this with Eq.~\eqref{eq:reverse-diffusion} and applying Girsanov theorem \cite{trick-1}, we obtain
\begin{eqnarray*}
\mathbb{E}_{p_{D}}\left[\log\frac{p_{D}(z_{[\cdot]})}{p_{E}(z_{[\cdot]} | x)}\right] 
&=& \mathbb{E}_{p_{D}}\left[\int_{0}^{T}g(t)\left(\nabla_{z_{t}}\log p_{D}(z_{t}) - s(z_{t},x,t;\theta)\right)\mathrm{d}\tilde{W}_{t}\right] \\
&& + \frac{1}{2}\mathbb{E}_{p_{D}}\left[\int_{0}^{T}g(t)^{2}\left\Vert \nabla_{z_{t}}\log p_{D}(z_{t}) - s(z_{t},x,t;\theta)\right\Vert^{2}\mathrm{d}t\right].
\end{eqnarray*}
Note that the first term on the right-hand side equals zero due to the martingale property of Itô integrals.  
Thus, we have
\footnotesize{\begin{eqnarray}
D_{\mathrm{KL}}\left(p_{D}(z_{[\cdot]},x)\,\|\,p_{E}(z_{[\cdot]},x)\right) 
&=& \mathbb{E}_{p_{D}}\left[\log\frac{p_{D}(x | z_{0})}{\pi(x)}\right] + \mathbb{E}_{p_{D}}\left[\log\frac{p_{D}(z_{[\cdot]})}{p_{E}(z_{[\cdot]} | x)}\right] \nonumber \\
&=& \mathbb{E}_{p_{D}}\left[\log\frac{p_{D}(x | z_{0})}{\pi(x)}\right] \nonumber \\
&& + \frac{1}{2}\mathbb{E}_{p_{D}}\left[\int_{0}^{T}g(t)^{2}\left\Vert \nabla_{z_{t}}\log p_{D}(z_{t}) - s(z_{t},x,t;\theta)\right\Vert^{2}\mathrm{d}t\right] \nonumber \\
&=& \mathbb{E}_{p_{D}}\left[\log\frac{p_{D}(x | z_{0})}{\pi(x)}\right] \nonumber \\
&& + \frac{1}{2}\mathbb{E}_{p_{D}}\left[\int_{0}^{T}g(t)^{2}\left(
\left\Vert \nabla_{z_{t}}\log p_{D}(z_{t})\right\Vert^{2}
+ \left\Vert s(z_{t},x,t;\theta)\right\Vert^{2}
\right)\mathrm{d}t\right] \nonumber \\
&& + \mathbb{E}_{p_{D}}\left[\int_{0}^{T}g(t)^{2}\left(s(z_{t},x,t;\theta)^{\top}\nabla_{z_{t}}\log p_{D}(z_{t})\right)\mathrm{d}t\right]. \label{eq:joint-kl-appendix}
\end{eqnarray}}

Moreover, we consider the following integration by parts:
\footnotesize{\begin{eqnarray*}
\mathbb{E}_{p_{D}}\left[s\left(z_{t},x\right)^{\top}\nabla_{z_{t}}\log p_{D}(z_{t}|z_{0})|x,z_{0}\right] & = & \int p_{D}(z_{t}|z_{0})s\left(z_{t},x\right)^{\top}\nabla_{z_{t}}\log p_{D}(z_{t}|z_{0})\mathrm{d}z_{t}\nonumber\\
 & = & \int s\left(z_{t},x\right)^{\top}\nabla_{z_{t}}p_{D}(z_{t}|z_{0})\mathrm{d}z_{t}\nonumber\\
 & = & \int\mathrm{div}_{z_{t}}\left(p_{D}(z_{t}|z_{0})s\left(z_{t},x\right)\right)\mathrm{d}z_{t} \nonumber\\
 & & -\int p_{D}(z_{t}|z_{0})\mathrm{tr}\left(\frac{\partial s(z_{t},x)}{\partial z_{t}}\right)\mathrm{d}z_{t}\nonumber\\
 & = & -\mathbb{E}_{p_{D}}\left[\mathrm{tr}\left(\frac{\partial s(z_{t},x)}{\partial z_{t}}\right)|z_{0}\right],\nonumber\\
\end{eqnarray*}}

\noindent{which yields}
\footnotesize{\begin{eqnarray}\mathbb{E}_{p_{D}}\left[s\left(z_{t},x\right)^{\top}\nabla_{z_{t}}\log p_{D}(z_{t}|z_{0})\right] & = & -\mathbb{E}_{p_{D}}\left[\mathrm{tr}\left(\frac{\partial s(z_{t},x)}{\partial z_{t}}\right)\right].\label{eq:integration-by-part}
\end{eqnarray}}

\noindent{By substituting Eq.~\eqref{eq:integration-by-part} into Eq.~\eqref{eq:joint-kl-appendix}, we finally obtain Eq.~\eqref{eq:edg-inequality-2}.}

It is worth noting that we could also directly design a loss function based on Eq.~\eqref{eq:joint-kl-appendix}.  
Such a loss is theoretically equivalent to the one used in this work, but may suffer from large variance in $\nabla_{z_{t}}\log p_{D}(z_{t})$ when $t$ is small.

\section{Proof of equivalence between Eq.~\eqref{eq:loss} and Eq.~\eqref{eq:loss-sa}}\label{append:proof-loss-sa}
First, we prove the Hutchinson estimator in Eq.~\eqref{eq:loss-sa}. In Eq.~\eqref{eq:loss}, one frequently needs the divergence $\nabla\cdot s(z_t, x, t;\theta)=\mathrm{tr}(\partial_{z_t}s)$ of a vector field $s: \mathbb{R}^d \rightarrow \mathbb{R}^d$. Deep learning frameworks (PyTorch, TensorFlow) do not natively provide an efficient way to form an entire $d\times d$ Jacobian for large $d$, while Hutchinson estimator replace $\mathrm{tr}(\cdot)$ with the unbiased Monte Carlo estimate without full Jacobians.

Considering a random variable $p(\epsilon)\sim\mathrm{Rademacher}^d$ with $\mathbb{E}[\epsilon]=0$ and $\mathrm{Cov}(\epsilon)=I$, we can get a stochastic estimation of the trace according to \cite{hutchinson1989stochastic}
\begin{eqnarray*}
    \mathrm{tr}(\partial_{z_t}s) &=& \mathbb{E}_{p(\epsilon)}[\epsilon^\top \partial_{z_t}s \ \epsilon].
\end{eqnarray*}
By the rules of matrix differential calculus,
\[
\mathrm{d}\bigl(\epsilon^\top s(z_t, x, t;\theta)\bigr) =\epsilon^\top\,\mathrm{d}s(z_t, x, t;\theta) =\epsilon^\top\,\bigl(\partial_{z_t} s(z_t, x, t;\theta)\bigr)\,\mathrm{d} z_t.
\]
Then, it enables the use of automatic differentiation for calculation,
\[
\epsilon^\top\,\bigl(\partial_{z_t} s(z_t, x, t;\theta)\bigr)\, = \frac{\partial \bigl[\epsilon^\top s(z_t, x, t;\theta)\bigr]}{\partial z_t} .
\]
Therefore, we have
\begin{eqnarray*}
    \nabla\cdot s(z_t, x, t;\theta) &=& \mathbb{E}_{p(\epsilon)}[\frac{\partial \bigl[\epsilon^\top s(z_t, x, t;\theta)\bigr]}{\partial z_t} \ \epsilon],
\end{eqnarray*}
where $p(\epsilon) \sim \mathrm{Rademacher}^d$ denotes a random vector in $\mathbb{R}^d$ with i.i.d.~Rademacher entries, satisfying $\mathbb{E}[\epsilon] = 0$ and $\mathrm{Cov}(\epsilon) = I$.

Then, we prove the expectation of a uniform time distribution $t$ over $[0, T]$ in Eq. \eqref{eq:loss-sa}. When $t\sim \mathcal{U}[0, T]$, its probability density function is
\[
p_{\mathcal{U}[0, T]}(t)= \begin{cases}\frac{1}{T}, & 0 \leq t \leq T \\ 0, & \text { otherwise }\end{cases} \ .
\]
Then we simplify the integration in Eq.~\eqref{eq:loss}
\begin{eqnarray*}
    \int_0^T\frac{g(t)^2}{2} \mathbb E_{p_D}[\cdot] \mathrm{d}t &=&\int_0^T \frac{1}{T} \frac{Tg(t)^2}{2} \mathbb E_{p_D}[\cdot] \mathrm{d}t \\
    &=&\int_{-\infty}^{+\infty} p_{\mathcal{U}[0, T]}(t) \frac{Tg(t)^2}{2} \mathbb E_{p_D}[\cdot] \mathrm{d}t \\
    &=&\mathbb E_{t\sim \mathcal{U}[0, T]}\left[ \frac{Tg(t)^2}{2} \mathbb E_{p_D}[\cdot]\right].
\end{eqnarray*}

Combining two parts above, we can achieve
\begin{eqnarray*}
    &&\int_0^T \frac{g(t)^2}{2} \mathbb{E}_{p_D}\left[\left\|s\left(z_t, x, t ; \theta\right)\right\|^2+2 \nabla_{z_t} \cdot s\left(z_t, x, t ; \theta\right)\right] \mathrm{d} t \\ 
    &=&\mathbb{E}_{t \sim \mathcal{U}[0, T], \epsilon \sim p(\epsilon),\left(x, z_t\right) \sim p_D\left(x, z_t\right)}\left[\frac{T g(t)^2}{2}\left(\left\|s\left(z_t, x, t ; \theta\right)\right\|^2+2 \frac{\partial\left[\epsilon^{\top} s\left(z_t, x, t ; \theta\right)\right]}{\partial z_t} \epsilon\right)\right],
\end{eqnarray*}
where $\mathcal{U}[0, T]$ denotes a uniform distribution over $[0, T]$, and $p(\epsilon)\sim \mathrm{Rademacher}^d$ denotes a random vector in $\mathbb{R}^d$ with i.i.d.~Rademacher entries, satisfying $\mathbb{E}[\epsilon] = 0$ and $\mathrm{Cov}(\epsilon) = I$.

\section{Calculation of the probability flow ODE}\label{append:ode}
Considering the SDE in Eq.~\eqref{eq:reverse-diffusion}, there is a corresponding deterministic ODE whose trajectories share the same marginal probability density
\[
\mathrm{d} z_t=\underbrace{\left\{f(z_t, t)-\frac{1}{2} g(t)^2 s_{\theta}(z_t, x, t)\right\}}_{=: F_{\theta}(z_t, x, t)} \mathrm{d} t .
\]
With the change of variables formula \cite{ode}, we can compute the log-likelihood of $p_E(z_0|x)$ using
\[
\log p_E(z_0|x) = \log p_E(z_T)+\int_0^T \nabla \cdot F_{\theta}(z_t, x, t) \mathrm{d} t.
\]
Due to the expensive computational cost of $\nabla \cdot F_{\theta}(z_t, x, t)$, the Hutchinson trace estimator technique in \ref{append:proof-loss-sa} is used again to calculate
\[
\nabla \cdot F_{\theta}(z_t, x, t)=\mathbb E_{p(\epsilon)}\left[ \frac{\partial[\epsilon^{\top} F_{\theta}(z_t, x, t)]}{\partial z_t} \epsilon\right],
\]
where $p(\epsilon) \sim \mathrm{Rademacher}^d$.

In our experiments, we employ the RK45 ODE solver implemented in \texttt{\seqsplit{scipy.integrate.solve\_ivp}}, where the parameters set identical to those in \citep{key-2} with \texttt{atol=1e-5} and \texttt{rtol=1e-5}.

\section{Analysis of Eq.~\eqref{eq:variational-logZ}}\label{append:logZ}
Based on Eq.~\eqref{eq:vae-loss} and $w(x,z_{0})=\frac{\exp(-U(x))p_{E}(z_{0}|x)}{p_{D}(z_{0})p_{D}(x|z_{0})}$, we have
\begin{eqnarray*}
\log Z &\ge& D_{\mathrm{KL}}(p_D(x)\|\pi(x))-\mathbb{E}_{p_{D}(z_{0}) p_{D}(x|z_{0};\phi)}\left[\log\frac{p_{D}(z_{0})p_{D}(x|z_{0};\phi)}{\exp(-U(x))p_{E}(z_{0}|x;\theta)}\right] \\
&=& D_{\mathrm{KL}}(p_D(x)\|\pi(x)) + \mathbb{E}_{p_{D}(z_{0}) p_{D}(x|z_{0};\phi)}\left[\log w(x,z_{0}) \right].
\end{eqnarray*}
Therefore, any Monte Carlo estimator of $\mathbb{E}_{p_{D}(z_{0}) p_{D}(x|z_{0};\phi)}\left[\log w(x,z_{0})\right]$ yields a lower bound on $\log Z$. 

Moreover, according to \ref{append:proof-vae-loss}, as the model parameters are optimized during training, the distribution $p_E(z_0|x;\theta)$ is expected to approximate the true conditional distribution $p_D(z_0|x)$ more closely. This improved approximation deduces 
\begin{eqnarray*}
    \mathbb{E}_{p_{D}(z_{0}) p_{D}(x|z_{0};\phi)}\left[\log w(x,z_{0}) \right] \rightarrow \log Z - D_{\mathrm{KL}}(p_D(x)\|\pi(x)).
\end{eqnarray*}
Consequently, a larger $\mathbb{E}_{p_{D}(z_{0}) p_{D}(x|z_{0};\phi)}\left[\log w(x,z_{0}) \right]$ suggests that the model is learning effectively, as $D_{\mathrm{KL}}(p_D(x)\|\pi(x))$ diminishes and the estimated $\log Z$ becomes more accurate.

\begin{algorithm}[b!]
  \caption{GHD based decoder}\label{alg:HM-transformation}
   \begin{algorithmic}[1]
   \Require $z_0=(\zeta,v_1,\ldots,v_K)$ drawn from the standard Gaussian distribution and decoder parameters $\phi$
   \State Let $y:=a(\zeta;\phi)$.
   \For{$k=1,\ldots,M$}
        \For{$j=1,\ldots,J$}
            \State Let $l:=\mathrm{leap\ frog\ step\ of\ GHD}$.
            \State \textcolor{black}{Update} $(y,v_i)$ by
\begin{eqnarray*}
v_{k} & := & v_{k}-\frac{\epsilon_0e^{\epsilon_0\epsilon(l;\phi)}}{2}\left(\nabla U(y)\odot e^{\frac{\epsilon_{0}}{2}Q_{v}(y,\nabla U(y),l;\phi)}+T_{v}(y,\nabla U(y),l;\phi)\right)\\
y & := & y+\epsilon_0 e^{\epsilon_0\epsilon(l;\phi)}\left(v_{k}\odot e^{\epsilon_{0}Q_{y}(v_{k},l;\phi)}+T_{y}(v_{k},l;\phi)\right)\\
v_{k} & := & v_{k}-\frac{\epsilon_0 e^{\epsilon_0\epsilon(l;\phi)}}{2}\left(\nabla U(y)\odot e^{\frac{\epsilon_{0}}{2}Q_{v}(y,\nabla U(y),l;\phi)}+T_{v}(y,\nabla U(y),l;\phi)\right)
\end{eqnarray*}
        \EndFor
        \State Let $v_k:=-v_k$
   \EndFor
   \State Let $\mu=y-\epsilon_0e^{\epsilon_0\eta(y;\phi)}\nabla U(y)$ and $\Sigma=2\epsilon_0e^{\epsilon_0\eta(y;\phi)}I$.
   \State \Return $x\sim \mathcal{N}(\mu,\Sigma)$, $p_D(x|z_0; \phi)$.
   \end{algorithmic}
\end{algorithm}

\section{Implementation of decoder}\label{append:hm-decoder}
The detailed implementation of the GHD based decoder is summarized in Algorithm \ref{alg:HM-transformation}. Within the algorithm, we initially stochastically generate a sample $y$ using the random variable $\zeta$ (refer to Line 1 of the algorithm). Subsequently, we iteratively update the sample using random velocities $v_1, \ldots, v_K$ and GHD (see Line 5). Finally, we utilize discretized Brownian dynamics with a trainable step size to design the decoder density of $x$ (refer to Line 9). Here, $\epsilon$, $Q_v$, $Q_y$, $T_v$, $T_y$ and $\eta$ are all neural networks composed of three-layer MLPs. $\epsilon_0$ serves as an initial hyperparameter of trainable step size $\epsilon$ and $\eta$, and our numerical experiments suggest that setting $\epsilon_0$ to a small positive value can improve the stability of the algorithm.

\section{Experimental details}\label{append:experimental-details}
\textbf{EDG } In our experiments, we leverage the subVP SDE proposed in \citep{key-2} to model the diffusion process of $z_t$ in Eq.~\eqref{eq:forward-sde}, defined as
\[
\mathrm{d} z_t = -\frac{1}{2}\beta(t) z_t \mathrm{d}t + \sqrt{\beta(t)\left(1-e^{-2\int_0^t \beta(s)\mathrm{d}s}\right)} \mathrm{d} W_t
\]
with $T=1$. Here, $\beta(t)=\beta_{\min}+t\left(\beta_{\max}-\beta_{\min}\right)$, and we set $\beta_{\min}=0.1$ and $\beta_{\min}=20$ for experiments.
The initial state $z_0$ follows a standard Gaussian distribution. Line 1 of Algorithm \ref{alg:HM-transformation} is implemented as
\[
y:=\mu_0(\zeta_0;\phi)+\Sigma_0(\zeta_0;\phi)\zeta_1,
\]
where $\zeta=(\zeta_0,\zeta_1)$, and $\mu_0(\zeta_0;\phi),\Sigma_0(\zeta_0;\phi)$ are modeled by MLPs.
In the GHD decoder design, we fix $J=5$ while setting $M=2$ for 2D energy function tasks, $M=10$ for Bayesian Logistic Regression tasks and Ising models, and $M=100$ for Lennard-Jones.

\textbf{Baseline}
For V-HMC, we facilitate the generation of samples after a warm-up phase comprising 1,000 steps, where each step the Metropolis-Hastings acceptance probability is computed to ensure the convergence.
For L2HMC, we use a three-layer fully connected network, setting the leapfrog step length to 10. For different tasks, we adaptively select varying step sizes.
For PIS, we use the publicly available code on our tasks directly.
Regarding BG, the RealNVP architecture consists of 3 affine blocks, where the scaling and translation functions are modeled by a three-layer fully connected network, each with 256 units and ReLU activation functions. 
\textcolor{black}{We would like to highlight that unlike data-driven energy-based models \cite{chao2023training,nijkamp2020mcmc}, our baseline BG generates samples solely from a prescribed energy function without any access to real data. The loss function is the KL divergence between the generated samples by BG and the target.}

\begin{table}[t!]
    \centerline{The model architecture of 2D energy tasks}
    \centering
    \small
    \resizebox{\textwidth}{!}
    {
    \begin{tabular}{ll}
\toprule
Decoding process & Encoding process \\
\midrule 
Network $a(\zeta;\phi)$: & Score network: \\
\ \ \ \ fc $10\times32$, ReLU, $2\times$(fc $32\times32$, ReLU), fc $32\times2$.& \ \ \ \ fc $13\times16$, ReLU,\\
GHD: & \ \ \ \ fc $16\times16$, ReLU,\\
\ \ \ \ $Q_v, T_v$: &   \ \ \ \ fc $16\times10$.  \\
\ \ \ \ \ \ \ \ fc $5\times10$, Tanh, fc $10\times10$, Tanh, fc $10\times2$.& \\
\ \ \ \ $Q_x, T_x$: & \\
\ \ \ \ \ \ \ \ fc $3\times10$, ReLU, fc $10\times10$, ReLU, fc $10\times2$.& \\
Final network $\textcolor{black}{\mathcal{N}}(\mu,\Sigma)$: & \\
\ \ \ \ fc $2\times10$, ReLU, fc $10\times10$, ReLU, fc $10\times2$. & \\
\bottomrule 
    \end{tabular}
}
\end{table}

\begin{figure}[t!]
\vspace{0mm}
\begin{center}
\includegraphics[width=\textwidth]{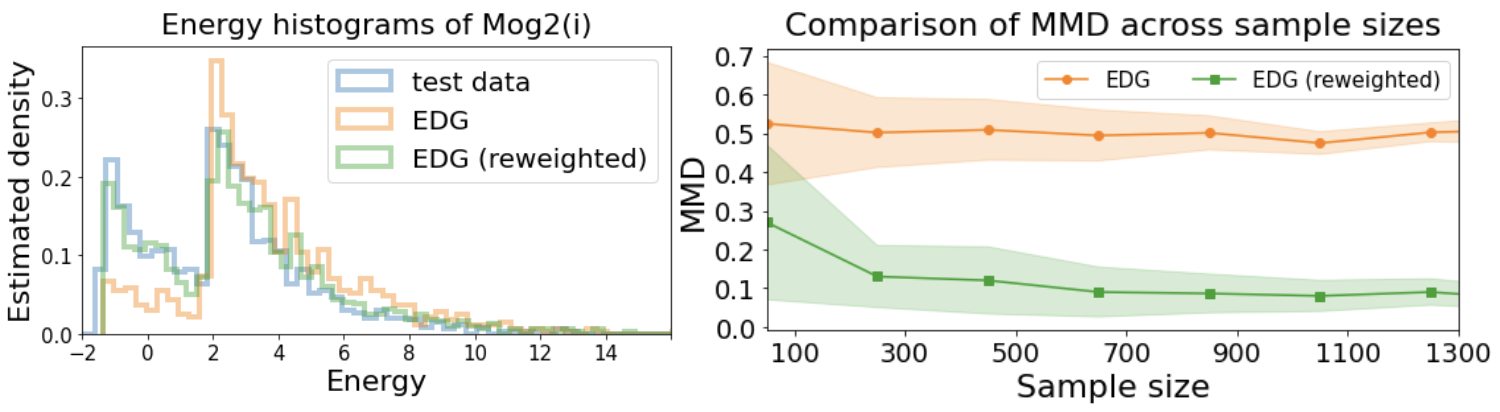}
\end{center}
\vspace{-7mm}
\caption{Reweighting effect in Mog2(i) task. \textbf{Left:} The energy histograms of $1000$ generated samples under EDG and EDG (reweighted). \textbf{Right:} The mean value of the reweighted MMD with respect to the sample size. We report the mean value $\pm$ standard deviation over 10 runs.}
\label{fig:mog2i-reweight-append}
\end{figure}

\textbf{2D energy function } We use the following model architectures to generate samples for 2D energy tasks. `fc $n$' denotes a fully connected layer with $n$ neurons. 

Reference samples for Mog2, Mog2(i), Mog6, and Mog9 in Fig.~\ref{fig:2d-example} are precisely drawn from the target distributions, as these distributions are mixtures of Gaussian distributions. Reference samples for Ring and Ring5 are generated using the Metropolis-Hastings algorithm, with a large number of iterations. Gaussian distributions with variances of 9 and 25, respectively, are used as the proposal distributions.
For the V-HMC, we employ the use of an HMC run of a total length of 2,000 steps, initiating from a standard normal distribution. The first 1,000 steps are designated as burn-in steps followed by a subsequent 1,000 steps used for sample generation. This procedure is independently executed 500 times for the purpose of creating the visual representation.

Given the reference samples $X=\{x_i\}_{i=1}^{m}$ with weights $w^X=\{w_i^X\}_{i=1}^{m}=\frac{1}{m}$ and the generated samples $Y=\{y_i\}_{i=1}^{m}$ with weights $w^Y=\{w_i^Y\}_{i=1}^{m}$, the Maximum Mean Discrepancy (MMD) \citep{mmd} measures the difference between the distributions of $X$ and $Y$ as follows:
{\footnotesize{\[
\mathrm{MMD}^2(X, Y) = \sum_i\sum_{j} w_i^Xw_j^X k(x_i, x_j) - 2\sum_i\sum_{j}w_i^Xw_j^Y k(x_i, y_j) + \sum_i\sum_{j}w_i^Yw_j^Yk(y_i, y_j),
\]}} 
\par\noindent where $k(\cdot, \cdot)$ denotes the kernel function to compute the inner product. RBF kernel is used in Table \ref{tab:result}, and the bandwidth is set as the median distance between corresponding samples. We conduct 10 independent runs to generate 5,000 samples with uniform density, and compute the mean MMD against 5,000 reference samples. The standard deviations are all small, hence we do not report them in the table.

\textcolor{black}{To better illustrate the effectiveness of the reweighting procedure in Sec. \ref{sec:marginal-encoder-density}, we use Mog2(i)—a case with room for further improvement—as an example. 
Firstly, we display the energy histograms of the generated samples under EDG and EDG (reweighted) in the left panel of Fig. \ref{fig:mog2i-reweight-append}.
Then, we compute the MMD using generated samples by EDG with normalized weights, which are calculated according to Sec. \ref{sec:marginal-encoder-density}. The mean value across 10 independent runs are reported in Table~\ref{tab:reweight_indicator-append}, each based on 5,000 samples. The standard deviations are omitted as they are negligibly small. The convergence of the weighted MMD with respect to the sample size is also presented in the right panel of Fig. \ref{fig:mog2i-reweight-append}.} \\

\begin{table}[t!]
 \vspace{-5mm}
    \small
    \centering
    \caption{The mean of the weighted MMD results over 10 rounds for EDG and EDG (reweighted), each using 5,000 samples. For EDG, all samples follow a uniform distribution. For EDG (reweighted), generated samples of EDG are assigned normalized weights, which are calculated according to Sec. \ref{sec:marginal-encoder-density}. The standard deviations are omitted as they are negligibly small.}\label{tab:reweight_indicator-append}
 \vspace{1mm}
    \begin{tabular}{ccc}
\toprule
 & EDG & EDG (reweighted) \\
Mean value of the reweighted MMD & $0.50$ & $0.09$ \\
\bottomrule 
    \end{tabular}
\vspace{0mm}
\end{table}

\textbf{Bayesian Logistic Regression} In all experiments, we employ the same data partition and the datasets are divided into training and test sets at a ratio of 4:1. Before training, we normalize all datasets to have zero mean and unit variance. For $d$-dimensional features, the architectures of neural networks involved in the EDG are as follows:

In the task of covertype, $x = (\alpha, w, b)$ with the prior distribution $p(x)=p(\alpha)p(w, b|\alpha)$, where $p(\alpha)=\mathrm{Gamma}(\alpha;1, 0.01)$ and $p(w, b|\alpha)=\mathcal{N}(w, b; 0,\alpha^{-1})$.

\begin{table}[h!]
    \centering
    \centerline{The model architecture of Bayesian Logistic Regression}
    \footnotesize
    \resizebox{\textwidth}{!}{
    \begin{tabular}{ll}
\toprule
Decoding process & Encoding process \\
\midrule 
Network $a(\zeta;\phi)$: & Score network: \\
\ \ \ \ fc $30\times 256$, ReLU, $2\times$(fc $256\times 256$, ReLU), fc $256\times d$.& \ \ \ \ fc $(d+31)\times 256$, ReLU,\\
GHD: & \ \ \ \ fc $256\times256$, ReLU,\\
\ \ \ \ $Q_v, T_v$: &   \ \ \ \ fc $256\times 30$.  \\
\ \ \ \ \ \ \ \ fc $(2d+1)\times 256$, Tanh, fc $256\times 256$, Tanh, fc $256\times d$.& \\
\ \ \ \ $Q_x, T_x$: & \\
\ \ \ \ \ \ \ \ fc $(d+1) \times 256$, ReLU, fc $256\times 256$, ReLU, fc $256\times d$.& \\
Final network $\textcolor{black}{\mathcal{N}}(\mu,\Sigma)$: & \\
\ \ \ \ fc $11d\times256$, ReLU, fc $256\times256$, ReLU, fc $256\times d$. & \\
\bottomrule 
    \end{tabular}
}
\end{table}


\begin{table}[b!]
    \centering
    \centerline{The model architecture of Lennard-Jones}
    \vspace{1mm}
    \footnotesize
    \resizebox{\textwidth}{!}{
    \begin{tabular}{ll}
\toprule
Decoding process & Encoding process \\
\midrule 
Network $a(\zeta;\phi)$: & Score network: \\
\ \ \ \ fc $100\times 32$, ReLU, $2\times$(fc $32\times 32$, ReLU), fc $32\times d$.& \ \ \ \ fc $(d+101)\times 256$, ReLU,\\
GHD: & \ \ \ \ fc $256\times256$, ReLU,\\
\ \ \ \ $Q_v, T_v$: &   \ \ \ \ fc $256\times 100$.  \\
\ \ \ \ \ \ \ \ fc $(2d+1)\times 32$, ReLU, fc $32\times 32$, ReLU, fc $32\times d$.& \\
\ \ \ \ $Q_x, T_x$: & \\
\ \ \ \ \ \ \ \ fc $(d+1) \times 32$, ReLU, fc $32\times 32$, ReLU, fc $32\times d$.& \\
Final network $\mathcal{N}(\mu,\Sigma)$: & \\
\ \ \ \ fc $d\times32$, ReLU, fc $32\times 32$, ReLU, fc $32\times d$. & \\
\bottomrule 
    \end{tabular}
}
\end{table}

\textcolor{black}{\textbf{Lennard-Jones} The LJ potential is an intermolecular potential which models repulsive and attractive interactions of non-bonding atoms or molecules. The energy is based on the distance of interacting particles $\mathcal{E}^{\mathrm{LJ}}(\cdot)$ and a harmonic potential $\mathcal{E}^{\mathrm{osc}}(\cdot)$ like \cite{klein2023equivariant}
\begin{eqnarray*}
    \mathcal{E}^{\mathrm{LJ}}(x) &=& \frac{1}{2} \sum_{i j}\left(\left(\frac{1}{d_{i j}}\right)^6-\left(\frac{1}{d_{i j}}\right)^{12}\right), \\
    \mathcal{E}^{\mathrm{osc}}(x) &=& \frac{1}{2} \sum_i\left\|x_i-x_{\mathrm{COM}}\right\|^2,
\end{eqnarray*}
where $d_{ij} = \| x_i - x_j \|^2$ is the Euclidean distance between particles $i$ and $j$ and $x_{\mathrm{COM}}$ refers to the center of mass of the system. The target energy function has the form $\mathcal{E}(\cdot) = \mathcal{E}^{\mathrm{LJ}}(\cdot) + \mathcal{E}^{\mathrm{osc}}(\cdot)$.
\textcolor{black}{The system is performed without periodic boundary conditions, which means particles are not wrapped across boundaries, no periodic images or minimum-image convention are used, and interactions are conducted within the finite simulation domain by harmonic confinement.}
For the experimental results, we evaluate the generated sample with the test sample from \cite{klein2023equivariant}. The test data were generated with MCMC with 1000 parallel chains, where each chain is run for 10,000 steps after a long burn-in phase of 200,000 steps starting from a random generated initial state. The density estimation results of interatomic distances for EDG and BG are plotted in Fig. \ref{fig:LJ-results}. The results for L2HMC and PIS are presented here in 
Fig. due to significant deviations from the test data. }

\begin{figure}[t!]
\begin{center}
\includegraphics[width=\textwidth]{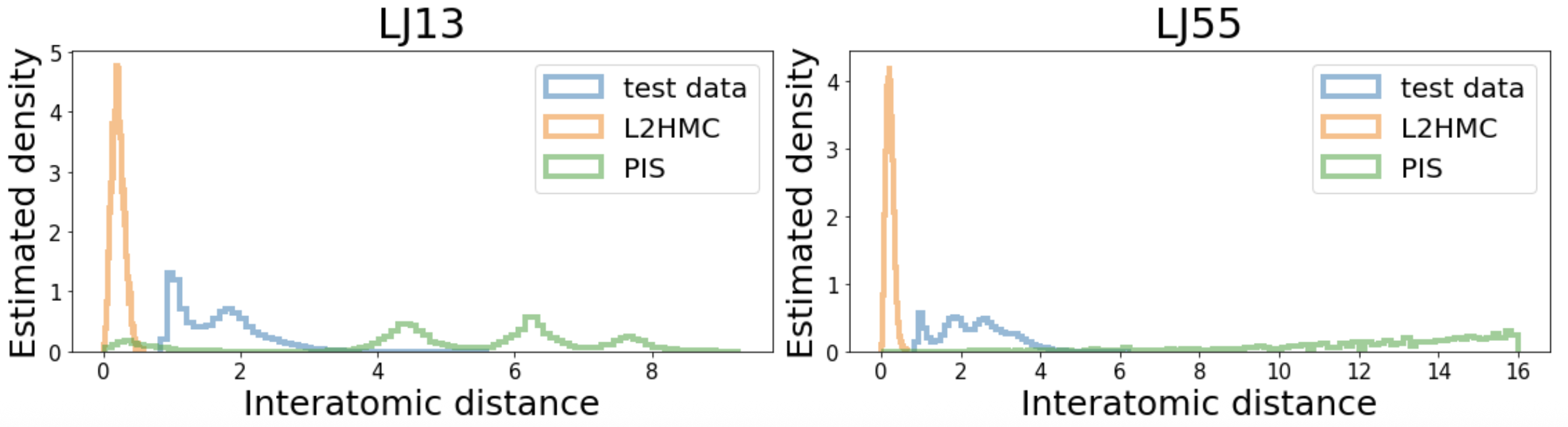}
\end{center}
\vspace{-7mm}
\caption{Comparison of the test data interatomic distance of LJ13 (left), LJ55 (right) with samples generated from L2HMC and PIS. Results for BG and EDG is ploted in Fig. \ref{fig:LJ-results}.}
\label{fig:computation}
\end{figure}

\begin{table}[b!]
\small
\begin{centering}
\vspace{-3mm}
\caption{\textcolor{black}{rESS} with mean $\pm$ standard deviation over 10 seeds. For each method, we use 256 samples for estimation.}\label{table:ess}
\vspace{1.5mm}
\begin{tabular}{cccc}
\toprule 
\textcolor{black}{rESS} & BG & PIS & EDG 
\tabularnewline
\midrule 
LJ13 & $0.006\pm0.002$ & $0.005\pm 0.001$ & $\mathbf{0.132\pm 0.048}$ \\
LJ55 & $0.004\pm 0.000$ & $0.004\pm 0.000$ & $\mathbf{0.098\pm 0.014}$
\tabularnewline
\bottomrule
\end{tabular}
\par\end{centering}
\end{table}

\textcolor{black}{
To evaluate the effectiveness of the sample reweighting strategy described in Sec.~\ref{sec:marginal-encoder-density}, we employ the relative Effective Sample Size (rESS), defined as
\begin{equation}\label{eq:ress}
\mathrm{rESS} = \frac{1}{N}\mathrm{ESS} = \frac{\left(\sum_{n=1}^N w(x^{n}, z_{0}^{n})\right)^2}{N\sum_{n=1}^N w(x^{n}, z_{0}^{n})^2}.
\end{equation}
Here, $\mathrm{ESS}$ denotes the Effective Sample Size \cite{freeman1966kish}, a widely used metric for evaluating the quality of weighted samples produced by importance sampling based on NFs (see, e.g., \cite{wirnsberger2023estimating, schebek2024efficient, jung2024normalizing, coretti2025learning}). The rESS value lies in the interval $(0, 1]$ and measures the fraction of effectively independent samples relative to the total sample size $N$. A low rESS indicates that the resulting estimators may suffer from high variance due to poor weight balance.
}

\textbf{Ising model } The model architecture is shown below, which is similar to that in NeuralRG \cite{ising} without stacking bijectors to form a reversible transformation. It retains the multiscale entanglement renormalization ansatz structure, while we only use one block to update the whole dimensions of the variable directly. To evaluate the efficiency of our proposed architecture, we include a comparative analysis against a standard MLP design in \ref{append:ablation}. The normalizing constant $\log Z$ is approximated according to Sec.~\ref{sec:marginal-encoder-density}.

\begin{table}[h!]
    \centering
    \centerline{The model architecture of Ising model}
    \footnotesize
    \begin{tabular}{ll}
\toprule
Decoding process & Encoding process \\
\midrule 
Network $a(\zeta;\phi)$: & Score network: \\
\ \ \ \ hierarchy network as shown in~\citep{ising}& \ \ \ \ hierarchy network. \\
GHD: & \ \ \ \ Input: $z_t$, $x$, $t$, $T$ \\
\ \ \ \ $Q_v, T_v$: &  \\
\ \ \ \ \ \ \ \ Conv2d $ 2 \times 10$, BatchNorm2d, ReLU, Conv2d $10\times 1$. & \\
\ \ \ \ $Q_x, T_x$: & \\
\ \ \ \ \ \ \ \ Conv2d $1 \times 10$, BatchNorm2d, ReLU, Conv2d $10\times 1$. & \\
Final network $\textcolor{black}{\mathcal{N}}(\mu,\Sigma)$: & \\
\ \ \ \ fc $d\times 10$, ReLU, fc $10 \times 10$, ReLU, fc $10 \times d$. & \\
\bottomrule 
    \end{tabular}
\end{table}

\section{Computation cost}\label{append:computation-cost}
\textcolor{black}{To demonstrate the computational efficiency of our model, we have compared the wall-clock training time and memory usage on the 2D energy (Mog2) task for the following models: the trainable MCMC model L2HMC \cite{mc-5}, the VI-based model BG \cite{key-4}, and the simulation-based model PIS \cite{diff-6}. For L2HMC and PIS, we adopted the default settings from their respective original papers. The network architectures for BG and EDG are detailed in \ref{append:experimental-details}. All experiments were conducted on a single NVIDIA RTX 2080 Ti GPU with 11GB of VRAM.  The runtime comparison is shown in Fig.~\ref{fig:computation}, and memory usage is reported in Table~\ref{tab:training_memory}.}

\begin{figure}[t!]
\begin{center}
\includegraphics[width=\textwidth]{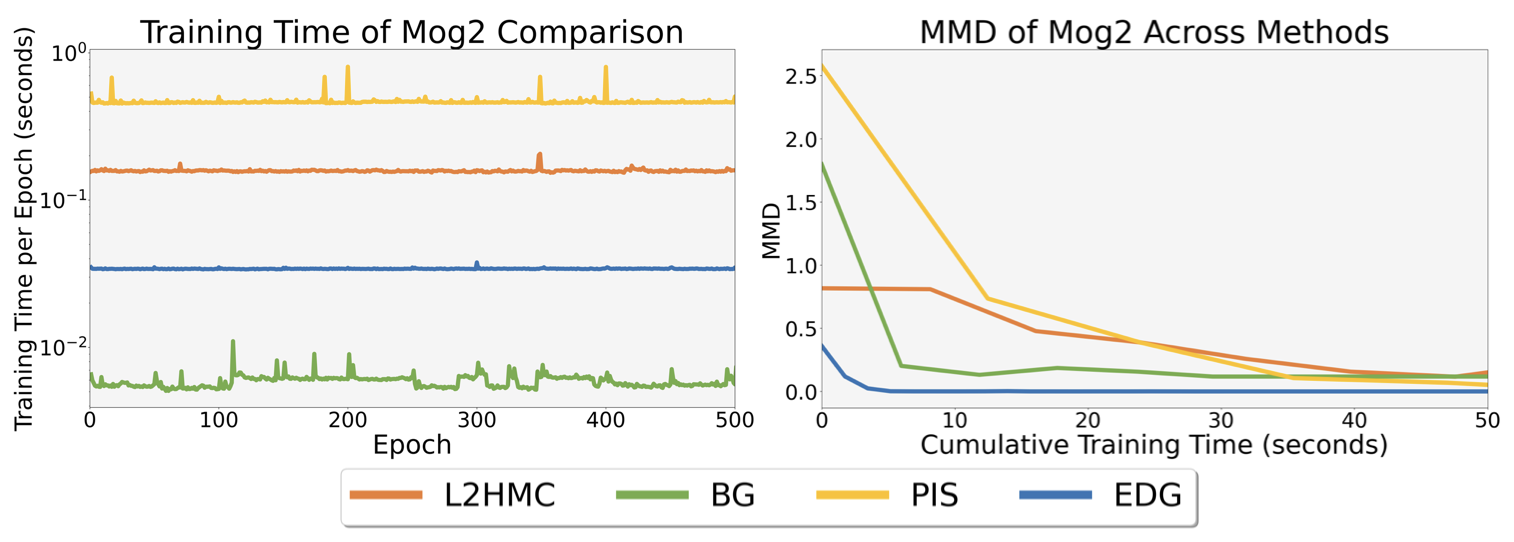}
\end{center}
\vspace{-7mm}
\caption{Computational time comparison across different methods. \textbf{Left:} the logarithm of the per-epoch training time for each method under a consistent batch size of 128. \textbf{Right:} the evolution of MMD in the Mog2 task with respect to cumulative training time, where all evaluation time has been excluded.}
\label{fig:computation}
\end{figure}

\begin{table*}[t!]
 \vspace{-5mm}
    \small
    \centering
    \caption{Training memory usage of different methods, measured by \texttt{torch.cuda.memory\_allocated}.}\label{tab:training_memory}
 \vspace{1mm}
    \begin{tabular}{ccccc}
\toprule
& L2HMC & BG & PIS & EDG  \\
\textbf{Memory (MB)} & 17.19 & 22.41 & 55.83 & 16.75  \\
\bottomrule 
    & & & & \\
    \end{tabular}
\vspace{-5mm}
\end{table*}

\begin{figure}[t!]
\begin{center}
\includegraphics[width=0.7\textwidth]{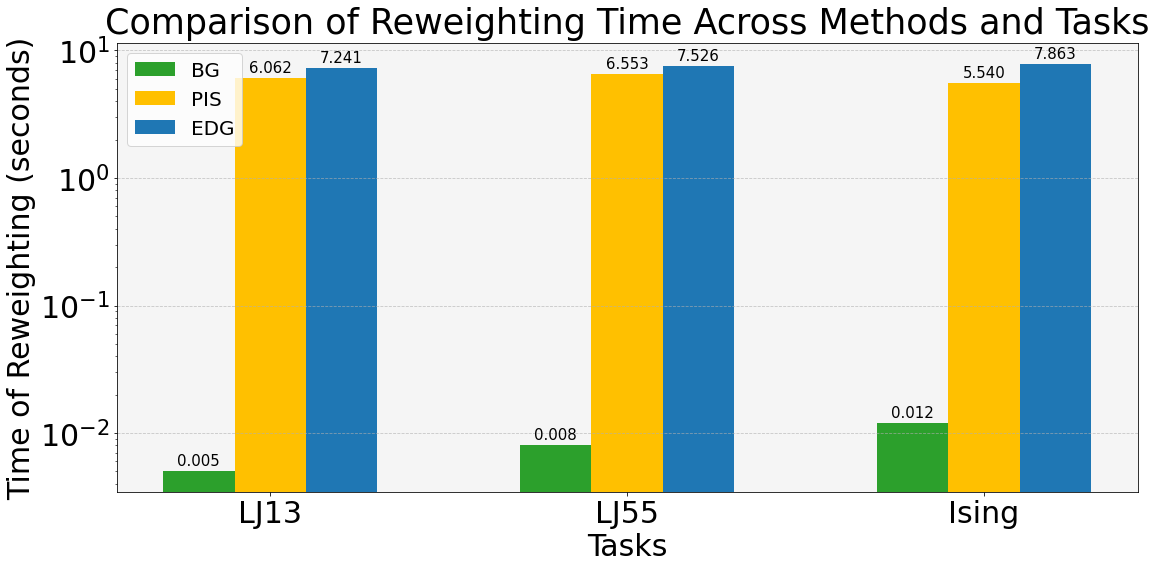}
\end{center}
\vspace{-8mm}
\caption{\textcolor{black}{The computational cost of sample reweighting in BG, PIS and EDG. Both PIS and EDG require temporal integration, for which distinct numerical schemes are employed to achieve comparable precision. PIS utilizes \texttt{torchsde.sdeint} with the Stochastic Runge-Kutta (SRK) method and a fixed step size of $T/1000$, while EDG adopts \texttt{scipy.integrate.solve\_ivp} using the RK45 method with \texttt{atol=1e-5}, \texttt{rtol=1e-5}.}}
\label{fig:reweighting_time}
\end{figure}

\textcolor{black}{The results show that EDG reaches competitive accuracy in less training time and with lower memory consumption compared to the other methods. Although the per-epoch training time of BG is shorter than that of EDG, EDG achieves better performance in fewer epochs.}

\textcolor{black}{
Additionally, we compared the reweighting time cost of three methods that support post-hoc sample reweighting: BG, PIS, and our proposed EDG. The results are presented in Fig.~\ref{fig:reweighting_time}.
Among them, BG achieves fast computation of the weight function due to the efficient evaluation of the Jacobian determinant in each affine coupling layer. In contrast, PIS and EDG must solve stochastic and ordinary differential equations respectively to compute weights, resulting in higher computational cost. The reweighting time of EDG and PIS is of the same order of magnitude, and EDG is slightly slower. This is mainly because the reweighting in EDG relies on the neural ODE technique, which involves divergence computation that contributes significantly to the time cost. PIS does not require divergence computation during reweighting.
It is worth emphasizing that PIS also requires numerical SDE solvers during training, while EDG avoids solving either SDEs or ODEs during training. As shown in Fig.~\ref{fig:computation}, this leads to significantly lower training time for EDG compared to PIS.}



\section{Ablation study}\label{append:ablation}
\textbf{GHD in Decoder:}
To elucidate the function of GHD in EDG, we compare the following models on the sampling task of 2D energy functions: a VAE with the Gaussian decoder and encoder, where the means and diagonal covariance matrices are both parameterized by MLPs; a VAE with a GHD-based decoder and MLP-based encoder; EDG without GHD, where the decoder is modeled by MLPs; and the full EDG model. For VAE w/o GHD, the network is composed of 3-layer MLPs, each with 32 units and ReLU activation function. Starting from $z\sim \mathcal{N}(0,I_d),\ d=10$, the decoder generates samples $x\sim p_D(x; \mu(z; \phi), \Sigma(z; \phi))$. The encoder, as an independent network, follows $p_E(z|x)=\mathcal{N}(z; \mu(x;\theta),\Sigma(x;\theta))$. The objective function is the KL divergence of the joint distribution of $z$ and $x$ as 
\[
D_{\mathrm{KL}}(p_D(z)p_D(x|z;\phi)||\pi(x)p_E(z|x;\theta).
\]
\par\noindent For VAE w/ GHD, the decoder is consistent with the structure in Algorithm \ref{alg:HM-transformation}, and the encoder is the same as described above. For EDG w/o GHD, we omit the leap-frog component (refer to Line 2-8 in the algorithm) and the decoder is composed of the network $a(\zeta;\phi)$ and a final gaussian part $\mathcal N(\mu,\Sigma)$ (refer to Line 1, 9 in the algorithm).

\begin{table}[h!]
 \vspace{-5mm}
    \small
    \centering
    \caption{The MMD between 5,000 samples generated by each generator and the reference samples.}\label{tab:ablation}
 \vspace{1mm}
    \begin{tabular}{ccccccc}
\toprule
& Mog2 & Mog2(i) & Mog6 & Mog9 & Ring & Ring5 \\
\midrule 
\textbf{VAE w/o GHD} & $1.86$ & $1.62$ & $2.57$ & $2.10$ & $0.12$ & $0.23$ \\
\textbf{VAE w/ GHD} & $\mathbf{0.01}$ & $2.43$ & $0.15$ & $0.59$ & $1.68$ & $0.63$ \\
\textbf{EDG w/o GHD} & $0.06$ & $1.01$ & $0.04$ & $0.05$ & $0.02$ & $0.04$ \\
\textbf{EDG} & $\mathbf{0.01}$ & $\mathbf{0.50}$ & $\mathbf{0.01}$ & $\mathbf{0.02}$ & $\mathbf{0.01}$ & \textcolor{black}{$\mathbf{0.02}$} \\
\bottomrule 
    & & & & &\\
    \end{tabular}
\vspace{-7mm}
\end{table}

\begin{figure*}[t!]
	\raggedright
    \vspace{-3mm}
    \subfigure{
    {\scriptsize{~~~~~~~~~\textbf{MoG2}~~~~~~~~~~~}}
    {\scriptsize{\textbf{MoG2(i)}~~~~~~~~~}}
    {\scriptsize{\textbf{MoG6}~~~~~~~~~~~}}
    {\scriptsize{\textbf{MoG9}~~~~~~~~~~~~}}
    {\scriptsize{\textbf{Ring}}~~~~~~~~~~~~}
    {\scriptsize{\textbf{Ring5}~~~~~~~~~~~}}
    }
    
        \vspace{-3mm}\hspace{-3.5mm}
	\subfigure{\rotatebox{90}{\scriptsize{~~~~~\textbf{VAE}}}
        \rotatebox{90}{\tiny{~~\textbf{w/o GHD}}}
		\begin{minipage}[t]{0.15\linewidth}
			\raggedright
			\includegraphics[width=1\linewidth]{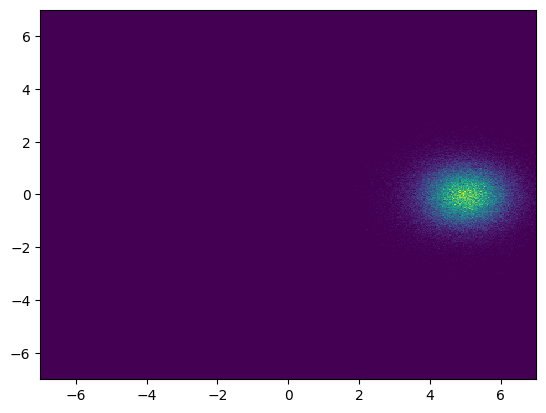}
		\end{minipage}
		\begin{minipage}[t]{0.15\linewidth}
			\raggedright
			\includegraphics[width=1\linewidth]{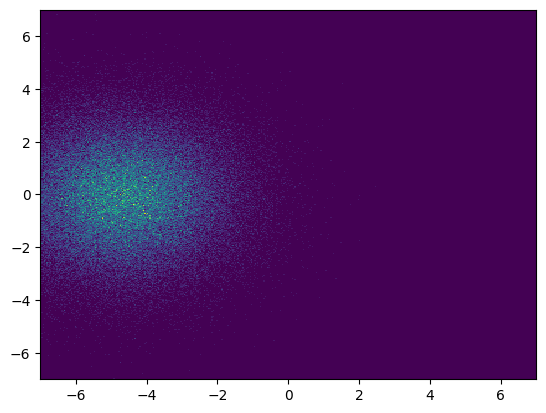}
		\end{minipage}
		\begin{minipage}[t]{0.15\linewidth}
			\raggedright
			\includegraphics[width=1\linewidth]{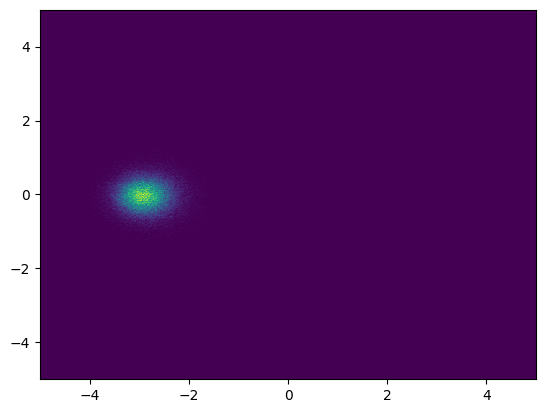}
		\end{minipage}
		\begin{minipage}[t]{0.15\linewidth}
			\raggedright
			\includegraphics[width=1\linewidth]{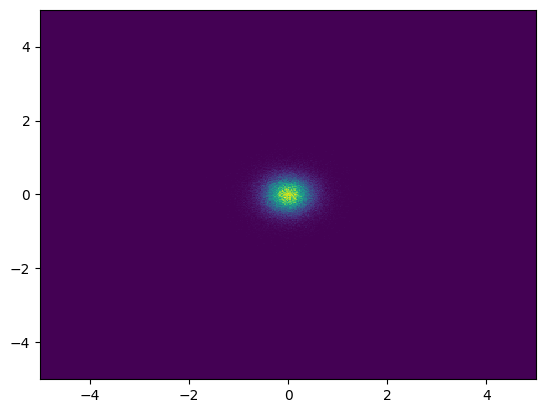}
		\end{minipage}
		\begin{minipage}[t]{0.15\linewidth}
			\raggedright
			\includegraphics[width=1\linewidth]{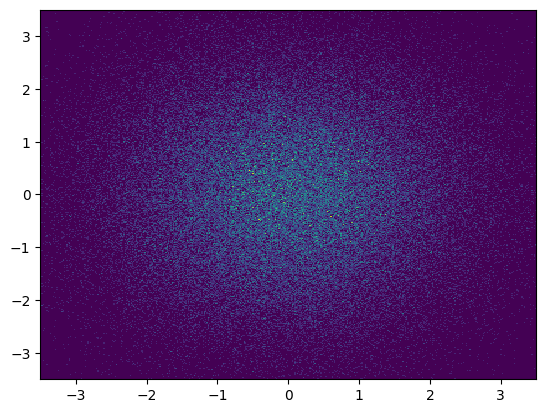}
		\end{minipage}
		\begin{minipage}[t]{0.15\linewidth}
			\raggedright
			\includegraphics[width=1\linewidth]{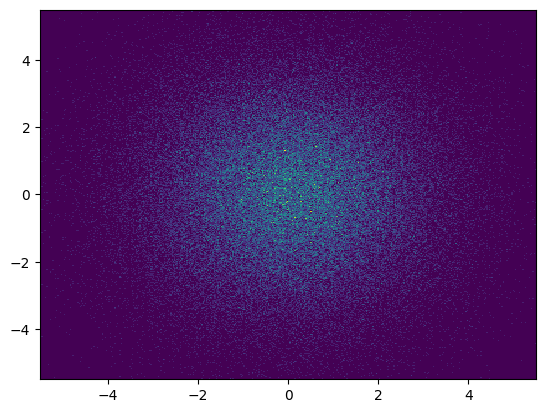}
		\end{minipage}
	}

    \vspace{-3mm}\hspace{-3.5mm}
	\subfigure{\rotatebox{90}{\scriptsize{~~~~~\textbf{VAE}}}
        \rotatebox{90}{\tiny{~~\textbf{w/ GHD}}}
		\begin{minipage}[t]{0.15\linewidth}
			\raggedright
			\includegraphics[width=1\linewidth]{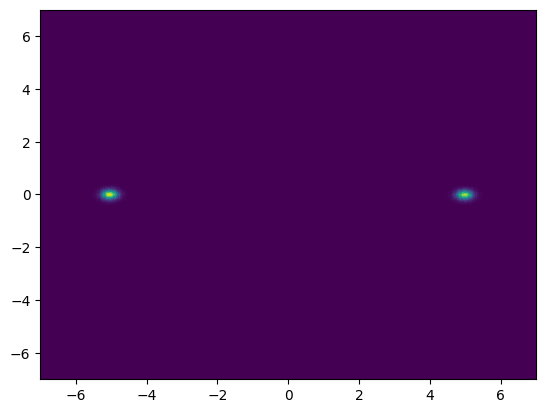}
		\end{minipage}
		\begin{minipage}[t]{0.15\linewidth}
			\raggedright
			\includegraphics[width=1\linewidth]{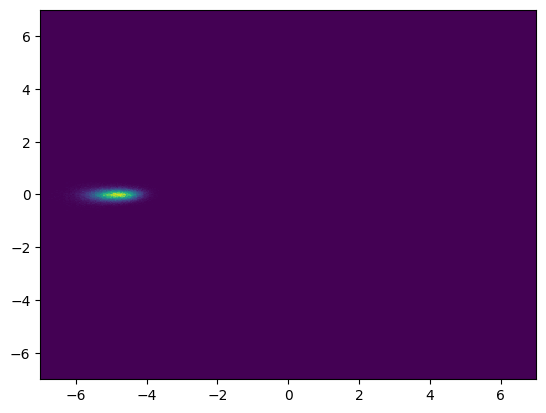}
		\end{minipage}
		\begin{minipage}[t]{0.15\linewidth}
			\raggedright
			\includegraphics[width=1\linewidth]{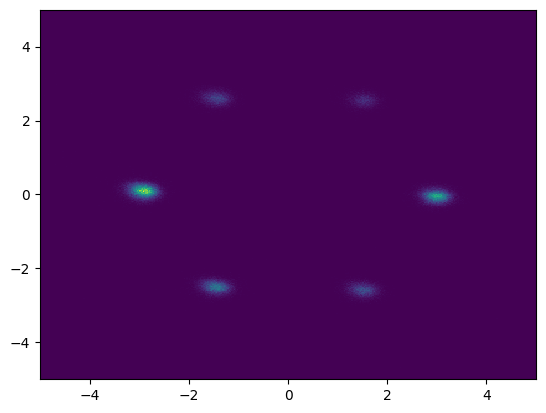}
		\end{minipage}
		\begin{minipage}[t]{0.15\linewidth}
			\raggedright
			\includegraphics[width=1\linewidth]{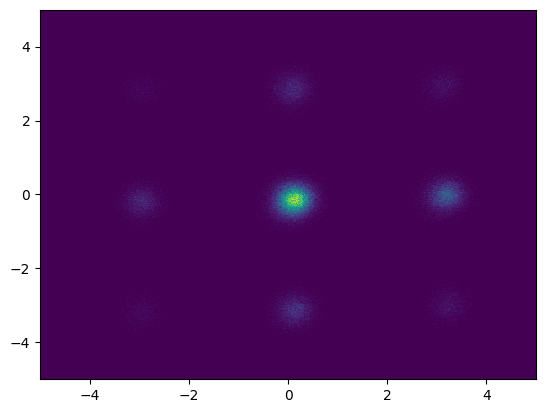}
		\end{minipage}
		\begin{minipage}[t]{0.15\linewidth}
			\raggedright
			\includegraphics[width=1\linewidth]{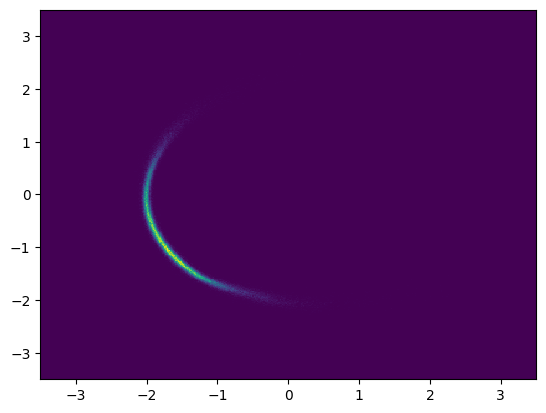}
		\end{minipage}
		\begin{minipage}[t]{0.15\linewidth}
			\raggedright
			\includegraphics[width=1\linewidth]{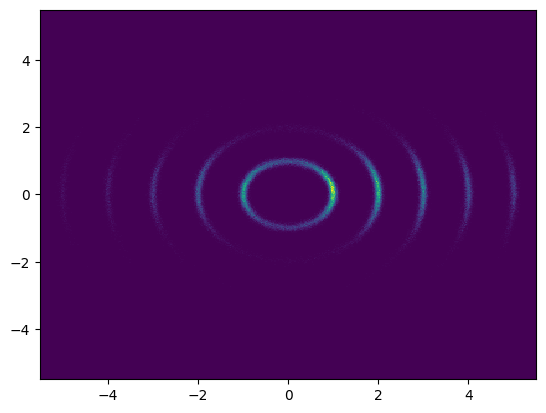}
		\end{minipage}
	}
    
    \vspace{-3mm}\hspace{-3.5mm}
	\subfigure{\rotatebox{90}{\scriptsize{~~~~~\textbf{EDG}}}
        \rotatebox{90}{\tiny{~~\textbf{w/o GHD}}}
		\begin{minipage}[t]{0.15\linewidth}
			\raggedright
			\includegraphics[width=1\linewidth]{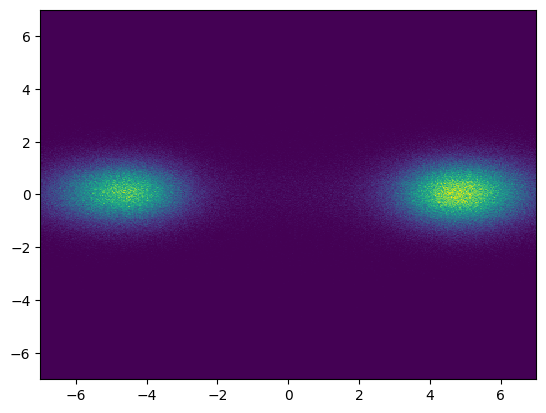}
		\end{minipage}
		\begin{minipage}[t]{0.15\linewidth}
			\raggedright
			\includegraphics[width=1\linewidth]{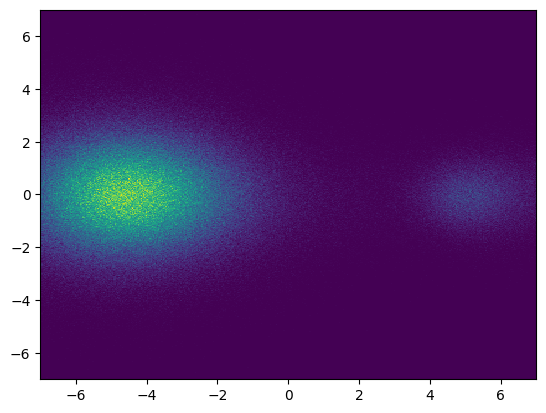}
		\end{minipage}
		\begin{minipage}[t]{0.15\linewidth}
			\raggedright
			\includegraphics[width=1\linewidth]{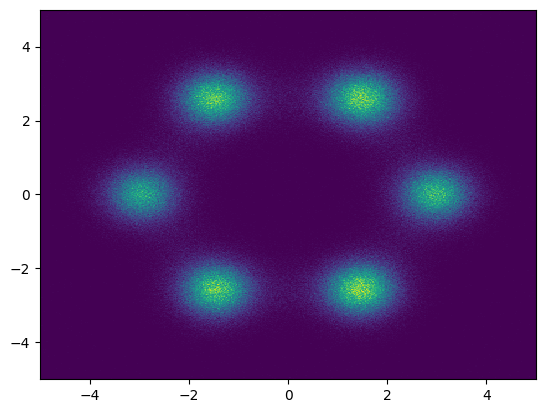}
		\end{minipage}
		\begin{minipage}[t]{0.15\linewidth}
			\raggedright
			\includegraphics[width=1\linewidth]{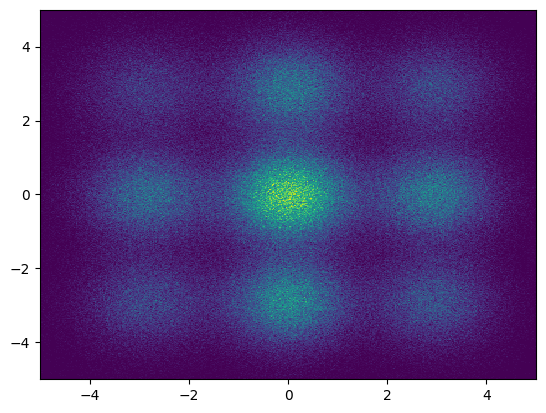}
		\end{minipage}
		\begin{minipage}[t]{0.15\linewidth}
			\raggedright
			\includegraphics[width=1\linewidth]{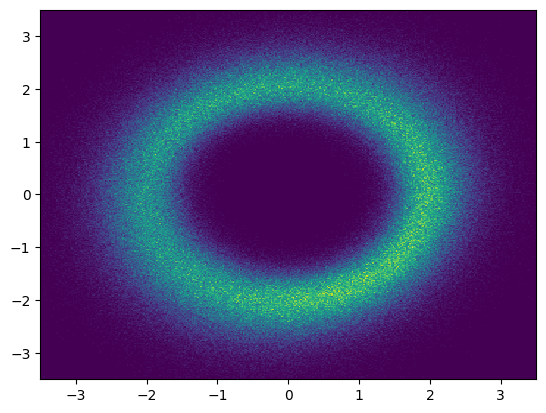}
		\end{minipage}
		\begin{minipage}[t]{0.15\linewidth}
			\raggedright
			\includegraphics[width=1\linewidth]{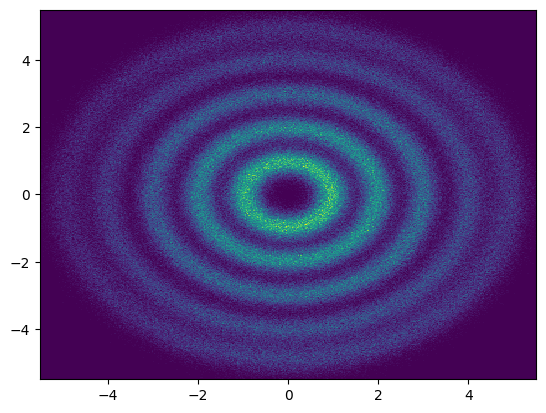}
		\end{minipage}
	}
 
	\vspace{-3mm}\hspace{-3.5mm}
	\subfigure{\rotatebox{90}{~~~~\scriptsize{\textbf{EDG}}}
        \rotatebox{90}{~~\tiny{\textbf{w/ GHD}}}
		\begin{minipage}[t]{0.15\linewidth}
			\raggedright
			\includegraphics[width=1\linewidth]{edg_hmc_Mog2_1.png}
		\end{minipage}
		\begin{minipage}[t]{0.15\linewidth}
			\raggedright
			\includegraphics[width=1\linewidth]{edg_hmc_Mog2_2.png}
		\end{minipage}
		\begin{minipage}[t]{0.15\linewidth}
			\raggedright
			\includegraphics[width=1\linewidth]{edg_hmc_Mog6.png}
		\end{minipage}
		\begin{minipage}[t]{0.15\linewidth}
			\raggedright
			\includegraphics[width=1\linewidth]{edg_hmc_Mog9.png}
		\end{minipage}
		\begin{minipage}[t]{0.15\linewidth}
			\raggedright
			\includegraphics[width=1\linewidth]{edg_hmc_ring.png}
		\end{minipage}
		\begin{minipage}[t]{0.15\linewidth}
			\raggedright
			\includegraphics[width=1\linewidth]{edg_hmc_ring5.png}
		\end{minipage}
	}
    \vspace{-8mm}
	\caption{Density plots for 2D energy function. We generate $500,000$ samples for each method and plot the histogram.}
	\label{fig:abaltion}
\end{figure*}

The histograms of the samples are shown in Fig.~\ref{fig:abaltion} for visual inspection and the sampling errors is summarized in Tab.~\ref{tab:ablation}. It is evident that EDG with a GHD-based decoder outperforms the other methods, demonstrating the effectiveness of each component within the model.

\textcolor{black}{\textbf{The step size of Decoder:} Step sizes will impact the model's performance for different tasks, selecting appropriate step sizes is crucial to achieve optimal results. To mitigate the influence of step size on the model's effectiveness, we treat them as trainable parameters with a hyperparameter $\epsilon_0$ in Sec.} \ref{sec:GHD}). 
\textcolor{black}{In our experiments, we systematically evaluated $\epsilon_0$ from 0.005 to 0.1 in increments of 0.005, selecting the optimal value based on training performance. The optimal value is provided in Tab.~\ref{tab:para}.} 

\begin{table}[h!]
 \vspace{-5mm}
    \small
    \centering
    \caption{Step size hyperparameters across tasks}\label{tab:para}
 \vspace{1mm}
     \begin{tabular}{ccccc}
\toprule
\textbf{Task} & 2d energy except Rings & 2d-Rings & & Bayesian Regression \\
\midrule 
\textbf{$\epsilon_0$} & 0.1 & 0.03 & & 0.05  \\
\bottomrule
\toprule
\textbf{Task} & Bayesian Regression-Cover & LJ13 & LJ15 & Ising model \\
\midrule 
\textbf{$\epsilon_0$} &  0.005 & 0.03 & 0.015 & 0.1 \\
\bottomrule 
    & & & \\
    \end{tabular}
\vspace{-7mm}
\end{table}

\begin{figure}[h!]
\begin{center}
\vspace{-3mm}
\includegraphics[width=0.7\textwidth]{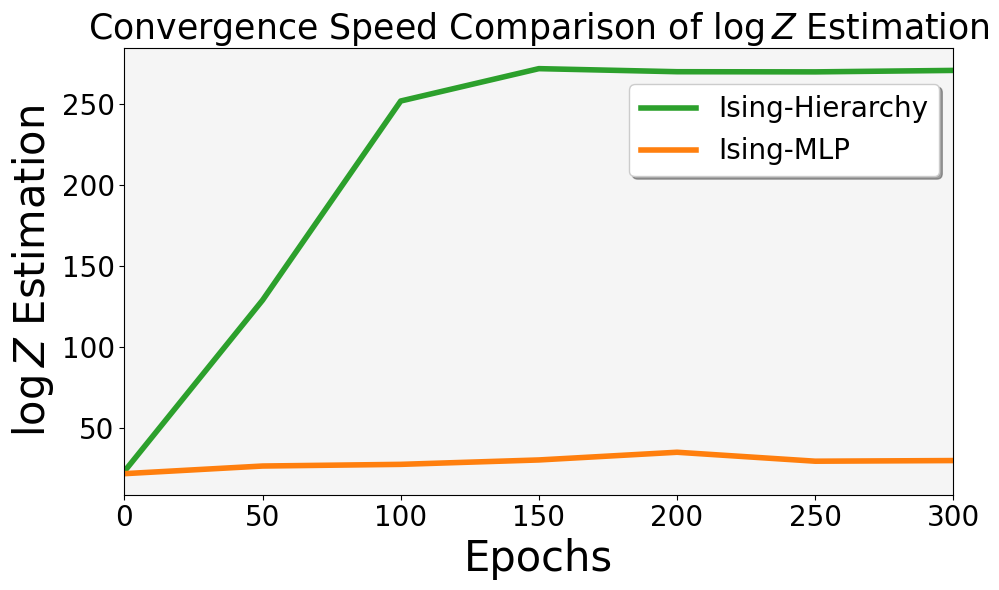}
\end{center}
\vspace{-7mm}
\caption{The $\log Z$ estimation curves under different model architectures in the Ising model task at $T=2.0$. Hierarchy denotes the model architecture in \ref{append:experimental-details} and MLP refers to EDG models where all networks are implemented as 3-layer MLPs with 32 hidden units and ReLU activation functions.}
\label{fig:nn_comparison_append}
\end{figure}

\textcolor{black}{\textbf{Network architecture:} To evaluate the effectiveness of the hierarchical design, we compare it with an EDG model using standard fully connected linear layers. As shown in Fig. \ref{fig:nn_comparison_append}, the hierarchical architecture better captures the underlying structure of the data and achieves superior performance. In future work, we plan to design specific network architectures for different tasks.} 

\end{document}